%% file: main.tex
\documentclass[final,5p,times,twocolumn]{elsarticle}

\usepackage{amsmath,amsfonts}
\usepackage{amsthm}
\usepackage{amssymb}
\usepackage{algorithmic}
\usepackage{algorithm}
\usepackage{array}
\usepackage[]{subfig}
\usepackage{textcomp}
\usepackage{stfloats}
\usepackage{url}
\usepackage{dutchcal}
\usepackage{verbatim}
\usepackage{graphicx}
\usepackage{tabularx}
\usepackage[table,xcdraw]{xcolor}
\usepackage{booktabs}
\usepackage[hidelinks]{hyperref}
\usepackage{cleveref}
\newtheorem{definition}{Definition}

\begin{document}

\begin{frontmatter}



\title{Dynamic Graph Representation Learning via Edge Temporal States Modeling and Structure-reinforced Transformer}


\author{
Shengxiang Hu$^a$,
Guobing Zou$^{a,}$*,
Song Yang$^a$,
Shiyi Lin$^a$,
Yanglan Gan$^b$,
Bofeng Zhang$^c$
} 

\affiliation{organization={School of Computer Engineering and Science, Shanghai University},
            city={Shanghai},
            postcode={200444}, 
            country={China}}

\affiliation{
    organization={School of Computer Science and Technology, Donghua University},
    city={Shanghai},
    postcode={201620}, 
    country={China}
}

\affiliation{
    organization={School of Computer and Information Engineering, Shanghai Polytechnic University},
    city={Shanghai},
    postcode={201209}, 
    country={China}
}

\begin{abstract}

Dynamic graph representation learning has emerged as a crucial research area, driven by the growing need for analyzing time-evolving graph data in real-world applications. While recent approaches leveraging recurrent neural networks (RNNs) and graph neural networks (GNNs) have shown promise, they often fail to adequately capture the impact of edge temporal states on inter-node relationships, consequently overlooking the dynamic changes in node features induced by these evolving relationships. Furthermore, these methods suffer from GNNs' inherent over-smoothing problem, which hinders the extraction of global structural features.
To address these challenges, we introduce the \underline{\textbf{R}}ecurrent \underline{\textbf{S}}tructure-reinforced \underline{\textbf{G}}raph \underline{\textbf{T}}ransformer (RSGT), a novel framework for dynamic graph representation learning. It first designs a heuristic method to explicitly model edge temporal states by employing different edge types and weights based on the differences between consecutive snapshots, thereby integrating varying edge temporal states into the graph's topological structure. We then propose a structure-reinforced graph transformer that captures temporal node representations encoding both graph topology and evolving dynamics through a recurrent learning paradigm, enabling the extraction of both local and global structural features.
Comprehensive experiments on four real-world datasets demonstrate RSGT's superior performance in discrete dynamic graph representation learning, consistently outperforming existing methods in dynamic link prediction tasks. 
\end{abstract}




\end{frontmatter}


\input{introduction}
\input{preliminaries}
\input{approach}
\input{experiments}
\input{related_work}

\section{Conclusion and Future Work}
\label{sec:conclusion}
This paper introduces a novel recurrent framework for dynamic graph representation learning, which we term as Recurrent Structure-reinforced Graph Transformer (RSGT).
In order to better understand the temporal dynamics of the graph, we propose to capture the influence of dynamic interactions between nodes on the strength of node relationships, thus we explicitly model the edge temporal states by converting each original snapshot into a weighted multi-relation graph according to the difference with the previous ones. Subsequently, we design a Structure-reinforced Graph Transformer (SGT) to recurrently learn and update node temporal representations across the dynamically modeled weighted multi-relation graphs.
The SGT is capable of effectively encapsulating global node semantic correlations, graph topology dependencies, and edge temporal state features concurrently, stems from the structure-aware attention-reinforced mechanism, which modifies the original node-wise attention score with pairwise topological structure and the corresponding shortest path. This leads to the generation of high-quality node representations that encode both global semantic relevance and structural information.
The effectiveness of the RSGT is validated through extensive experiments, which exhibit superior performance in understanding the evolutionary characteristics of a dynamic graph compared to the existing baseline methods. 

In future work, we plan to extend RSGT to heterogeneous dynamic graphs and more complex application scenarios. Real-world networks often involve multiple types of nodes and edges with diverse attributes, which can provide richer contextual information. Adapting our model to these heterogeneous environments would enable more nuanced modeling of complex systems across various domains such as social networks, e-commerce, and biomedical research. This extension could potentially unlock new insights and applications in dynamic graph analysis, further demonstrating the versatility and power of our proposed approach.

\section*{Acknowledgments}
This work was supported by National Natural Science Foundation of China (No. 62272290, 62172088), and Shanghai Natural Science Foundation (No. 21ZR1400400).

 \bibliographystyle{elsarticle-num} 
 \bibliography{ref}






\end{document}

%% file: introduction.tex
\section{Introduction}
Graphs find extensive use across myriad application domains, such as traffic flow forecasting \cite{10184800,DBLP:journals/tits/DuCLCL24}, recommender systems \cite{DBLP:journals/tois/YiOM24, DBLP:journals/tkde/YuXCCHY24, DBLP:journals/tkde/LiuWWYDZW24} and stock prediction \cite{10176355}, to delineate the interactions between entities within real-world complex systems.
As these systems inherently exhibit temporal variations, with entities and interactions frequently emerging or disappearing, they can be intuitively abstracted as dynamic graphs. Dynamic graphs, which capture the evolution of relationships over time, can be categorized as either discrete \cite{pareja2020evolvegcn,you2022roland} or continuous \cite{nguyen2018continuous,wen2022trend} based on how they model temporal changes.
For instance, in social network analysis, a dynamic graph can represent the evolving patterns of user interactions, where edges appear or disappear as users form new connections or cease communication.
Learning the representation of these dynamic graphs is of paramount importance as it encapsulates the interaction and evolution mechanisms of the graph, thereby aiding in understanding and predicting the system behavior.

In contrast to static graph representation learning \cite{perozzi2014deepwalk,grover2016node2vec}, which focuses primarily on capturing fixed topological features, dynamic graph representation learning faces the dual challenge of preserving evolving topological structures while simultaneously capturing the temporal dynamics of the graph. This dual requirement significantly increases the complexity of model design, necessitating architectures that can effectively capture both spatial dependencies and temporal evolution patterns concurrently. The key challenge lies in maintaining the structural integrity of the graph at each time step while effectively modeling the temporal evolution of both node attributes and edge formations. In this study, we focus primarily on discrete dynamic graph representation learning. This choice is motivated by its widespread applicability across various domains \cite{pareja2020evolvegcn,wang2021modeling,you2022roland} and its efficacy in capturing significant state changes in real-world scenarios, where graph structures often evolve in distinct, observable steps rather than continuous transitions.

A discrete dynamic graph is modeled as a sequence of ordered static snapshots across discrete time intervals, capturing changes in graph topology, including node and edge modifications. Traditional approaches \cite{pareja2020evolvegcn,wang2021modeling,hu22temporal} have combined Graph Neural Networks (GNNs) with sequence modeling techniques to capture both structural and temporal information. Recent research has explored more sophisticated architectures, such as transformers and advanced recurrent structures \cite{wang2021modeling}, to better handle long-range dependencies and improve inductive learning capabilities.
In parallel, continuous dynamic graph modeling methods have gained traction, treating graph evolution as a series of timestamped events. These approaches, exemplified by works like TGAT \cite{DBLP:conf/iclr/XuRKKA20} and CAW \cite{DBLP:conf/iclr/WangCLL021}, leverage attention mechanisms and temporal random walks to capture fine-grained temporal-topological interactions. Advanced techniques such as neural ordinary differential equations \cite{yao2024recurrent} and stochastic processes \cite{wen2022trend,celikkanat2024continuous} have been employed to model continuous-time dynamics more precisely.
Emerging research trends focus on developing methods that can effectively balance the modeling of both short-term and long-term temporal dependencies \cite{pareja2020evolvegcn,goyal2018dyngem}, while also addressing the challenges of scalability and computational efficiency in large-scale dynamic graphs \cite{wang2024contig}. These advancements aim to provide more comprehensive and nuanced representations of dynamic graph structures, capturing both local and global patterns of evolution.

However, existing approaches exhibit limitations that constrain their effectiveness in dynamic graph representation learning. Firstly, current methods often overlook the dynamic influence of edge temporal states on node features. The temporal evolution of edges—their persistence, emergence, or disappearance—significantly impacts node characteristics, yet many approaches fail to explicitly model these dynamics \cite{zhu2023path,zhou2022point}. Secondly, while recent research \cite{ying2021transformers,chen2022structure} has begun to explore transformer architectures to address the over-smoothing problem in GNNs \cite{chen2020measuring, rusch2023survey}, challenges remain in effectively integrating graph structural information with complex temporal dependencies. Moreover, many current approaches struggle to leverage historical node information inductively, limiting their applicability to evolving networks. Consequently, there is a need for an approach that can adaptively integrate temporal and structural information while maintaining generalizability to unseen nodes and edges.

To address the aforementioned challenges, we propose a novel framework for dynamic graph representation learning, named \underline{\textbf{R}}ecurrent \underline{\textbf{S}}tructure-reinforced \underline{\textbf{G}}raph \underline{\textbf{T}}ransformer (RSGT). 
At the core of RSGT is a unique graph transformation technique that converts each graph snapshot into a weighted multi-relation graph. This transformation explicitly models edge temporal states through distinct edge types and weights, capturing the variations between consecutive snapshots. 
Next, we propose a structure-reinforced graph Transformer that introduces a novel structure-aware attention mechanism, modifying the original attention computation to make the Transformer inherently cognizant of graph structures. By operating on our weighted multi-relation difference graph, this enhanced architecture enables the model to simultaneously capture both intricate topological features and temporal dynamics. The structure-aware attention mechanism selectively focuses on salient historical patterns and structural relationships encoded in edge types and weights, effectively learning the complex interplay between graph evolution and temporal dependencies. 
We then conceptualize dynamic graph representation learning as a sequence of recurrent learning tasks. Starting with initial node features, RSGT continuously updates these representations based on the evolving graph structure and edge states, enabling RSGT to effectively capture long-term dependencies in the dynamic graph.
Moreover, the recurrent learning paradigm allows for incremental updates of node representations rather than recomputing from scratch at each time step.

Our extensive empirical evaluations on dynamic link prediction tasks across four real-world datasets demonstrate RSGT's superior performance in dynamic graph representation learning, highlighting its potential for dynamic graph analysis. 
In conclusion, the key contributions of this paper can be summarized as follows:
\begin{itemize}
    \item We propose RSGT, a novel recurrent learning framework for dynamic graphs. By transforming graph snapshots into weighted multi-relational difference graphs, RSGT captures edge temporal states and integrates them with topological structures, enabling a more comprehensive representation of dynamic graph evolution.
    \item We design a structure-reinforced graph transformer that simultaneously extracts local-global topological features and evolving dynamic characteristics, effectively mitigating the over-smoothing problem while enhancing the model's ability to capture complex temporal-structural dependencies.
    \item Extensive experiments on real-world datasets demonstrate RSGT's superior performance in dynamic link prediction tasks. The results not only validate the effectiveness of our approach in capturing intricate dynamic graph patterns but also highlight its potential for advancing the field of dynamic graph analysis.
\end{itemize}

The remainder of this paper is organized as follows: Section \ref{sec:formulation} provides preliminaries and problem formulation. Section \ref{sec:approach} details the proposed RSGT framework. Experimental results are presented in Section \ref{sec:exp}. Related work is reviewed in Section \ref{sec:related}, followed by conclusions in Section \ref{sec:conclusion}.

%% file: preliminaries.tex
\section{Problem Formulation}
\label{sec:formulation}


Here, we present the formulations for dynamic graph representation learning through a set of definitions. 

\begin{definition}[Dynamic Graph]
    A discrete dynamic graph, denoted as $\mathcal{G}$, is an ordered sequence of static graphs represented as $\{\mathcal{G}_1, \dots, \mathcal{G}_t\}$. Here, $\mathcal{G}_t = \langle V_t, E_t\rangle$ corresponds to a snapshot of the dynamic graph $\mathcal{G}$ at time slice $t$, where $V_t$ is the set of nodes, $E_t$ is the set of edges. An edge $e_{ij}^t \in E_t$ indicates a link between nodes $v_i, v_j \in V_t$ at time $t$.
\end{definition}

While the dynamic graph captures the overall structure at each time step, it's crucial to understand how individual edges evolve over time. This leads us to the concept of edge temporal states:

\begin{definition}[Edge Temporal States]
    The edge temporal states, denoted as $S_t$, capture the dynamic nature of edges in a graph over time. For each edge $e_{ij}^t \in E_t$, its temporal state $s_{ij}^t \in S_t$ is defined as a tuple $s_{ij}^t = \langle tp_{ij}^t, \omega_{ij}^t \rangle$. Here, $tp_{ij}^t$ represents the current state of the edge, which can be persistent (the edge exists at both $t-1$ and $t$), emerging (the edge appears at $t$ but did not exist at $t-1$), or disappearing (the edge existed at $t-1$ but no longer exists at $t$). The second component, $\omega_{ij}^t$, represents the duration of the current state, indicating the number of consecutive time slices the edge has been in its current state.
\end{definition}

\begin{figure*}[t]
    \centering
     \includegraphics[width=\textwidth]{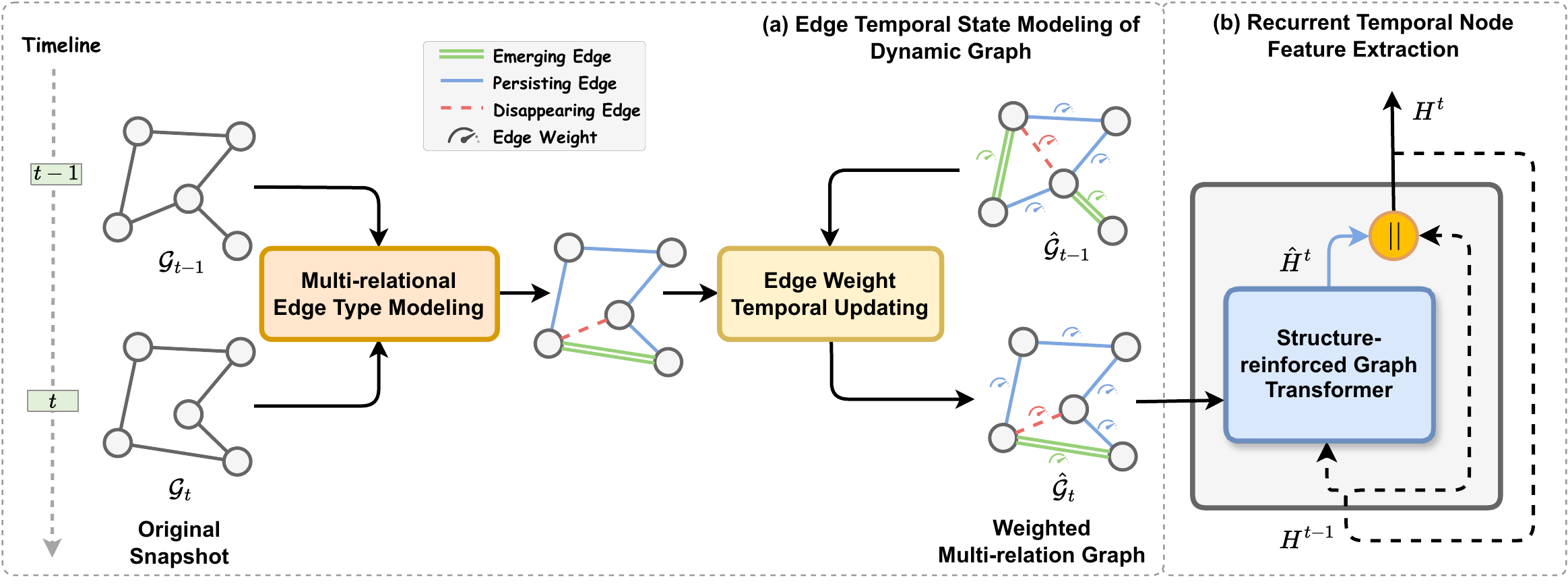}
    \caption{Overall framework of the proposed Recurrent Structure-reinforced Graph Transformer (RSGT). (a) Edge Temporal State Modeling of Dynamic Graph: transforming each graph snapshot into a weighted multi-relation graph to model the edge temporal states. (b) Recurrent Temporal Node Feature Extraction: capturing both graph topology and evolving dynamics through a recurrent learning paradigm with the structure-reinforced graph transformer.}
    \label{fig:framework}
\end{figure*}

While dynamic graphs and edge temporal states provide a framework for describing evolving systems, effectively analyzing these systems requires learning compact representations that capture both structural and temporal information. This leads us to the task of dynamic graph representation learning:

\begin{definition}[Dynamic Graph Representation Learning]
    Dynamic graph representation learning aims to learn a mapping function that derives a low-dimensional representation $\textbf{H}^t\in \Re^{|V_t|\times d}$ ($d\ll |V_t|$) by utilizing a succession of snapshots $\{\mathcal{G}_1, \dots, \mathcal{G}_{t}\}$ and their corresponding edge temporal states $\{S_1, \dots, S_t\}$ up to time slice $t$.
    The $i$-th row of $\textbf{H}^t$, denoted as $\textbf{h}^t_{i}$, represents the derived representation of node $v_i$ at time slice $t$, encapsulating both structural information and evolutionary dynamics.
    A typical learning framework can be formulated as:
    \begin{equation}
    \textbf{H}^t = g(f(\mathcal{G}_t, S_t, \textbf{X}|\Theta_f), \textbf{I}_{t-1}|\Theta_g)
    \label{equ:repre}
    \end{equation}
    where $\textbf{X}$ denotes the initial node features or embedding matrix, $\textbf{I}_{t-1}$ represents the most recent latent temporal state, $f$ is an encoding model for extracting structural and temporal features, and $g$ is a sequence model for discerning evolving patterns. $\Theta_f$ and $\Theta_g$ are the learnable parameters of $f$ and $g$ respectively.
\end{definition}

The effectiveness of node representations learned through this process is crucial for various graph-oriented tasks \cite{quach2021dyglip,wen2022trend,xu2019spatio}. One particularly important application is dynamic link prediction, which we use in this study to evaluate our proposed method:

\begin{definition}[Dynamic Link Prediction]
    Given a series of snapshots $\{\mathcal{G}_1, \dots, \mathcal{G}_t\}$ up to time slice $t$, the dynamic link prediction task aims to predict whether a candidate edge $e_{ij}^{t+1}$ will exist in the upcoming snapshot $\mathcal{G}_{t+1}$ at time slice $t+1$. This can be formulated as:
    \begin{equation}
        p_{ij}^{t+1} = \tau(\mathcal{G}_{t+1}|\mathcal{G}_1, \dots, \mathcal{G}_t, S_1, \dots, S_t;\Theta_{\tau})
    \end{equation}
    where $p_{ij}^{t+1}$ represents the predicted probability that $e_{ij} \in E_{t+1}$, and $\tau$ indicates the prediction model with parameters $\Theta_{\tau}$.
\end{definition}

%% file: approach.tex
\section{Approach}
\label{sec:approach}

The overall framework of our proposed RSGT is depicted in Figure \ref{fig:framework}. It consists of two essential modules: 
(a) Edge Temporal State Modeling of Dynamic Graph: This module transforms each original graph snapshot into a weighted multi-relation difference graph. It assigns diverse relational types and weights to edges based on their temporal evolution between adjacent snapshots. Specifically, it employs a multi-relational edge type modeling technique to capture short-term dynamics and an edge weight temporal updating mechanism to represent long-term dynamics. 
(b) Recurrent Temporal Node Feature Extraction: In this module, we introduce a structure-reinforced graph transformer that simultaneously considers the graph topology and the edge temporal states over the weighted multi-relation difference graph. This transformer is used to recurrently update node features, effectively capturing both local structural information and global temporal dependencies.
These two modules work in tandem to learn comprehensive and time-aware node representations, enabling RSGT to effectively model the complex dynamics of evolving graphs. In the subsequent sections, we delve deeper into the intricacies of each module.

\subsection{Edge Temporal State Modeling of Dynamic Graph}

In a dynamic graph $\mathcal{G}$, the feature of a focal node $v_i$ can be intuitively perceived as a reflection of its interactions with other nodes $v_j \in (N_i^{h})^t$ that it directly or indirectly engages with \cite{wang2019neural}. Here, $(N_i^{h})^t$ signifies the $h$-hop neighborhood of $v_i$ at time slice $t$, representing all nodes that can be reached from $v_i$ within $h$ steps at time $t$. 

Social network theory, as observed in \cite{zhu2023path,zhou2022point}, suggests that the strength of relationships between nodes predominantly depends on factors such as historical interaction frequency and duration. This relationship strength, in turn, shapes future interactions, thus playing a significant role in the formation and evolution of the graph. However, capturing these dynamic relationships in a way that effectively informs node representation learning presents several challenges. The interactions between nodes exhibit persistence, emergence, or disappearance across multiple time slices, resulting in perpetual alterations in the temporal states of the edges. Moreover, the temporal state of a specific edge is primarily characterized by two features: short-term dynamics, which represent the edge's immediate state changes, and long-term dynamics, which constitute the aggregate impact of this edge on the strength of the correlation between its associated nodes. Understanding the evolution of $\mathcal{G}$ thus hinges on accurately learning how distinct edge temporal states impact node features across different time slice snapshots.

To address these challenges and capture these complex dynamics, we propose explicitly modeling edge temporal states, which can provide heuristic information to the downstream feature extraction model, thereby enabling effective learning of node representations. To this end, we propose a heuristic edge temporal state modeling method that transform each $\mathcal{G}_t \in \mathcal{G}$ into a weighted multi-relation difference graph $\hat{\mathcal{G}}_t$ based on the differences between adjacent graph snapshots. This graph uses different edge types and edge weights to explicitly model the temporal states of edges, capturing both short-term and long-term dynamics. Formally, we define this graph as follows:

\begin{definition}[Weighted Multi-relation Difference Graph]
A weighted multi-relation difference graph $\hat{\mathcal{G}}_t = \langle V_t, \hat{E}_t, S_t \rangle$ at time slice $t$ is defined as a graph comprising a set of nodes $V_t$, a set of edges $\hat{E}_t$ encompassing connections from both current and previous time slices, and a set of edge temporal states $S_t=\{\langle tp_{ij}^t, \omega_{ij}^t \rangle \}$, where $tp_{ij}^t$ represents the edge type capturing short-term dynamics, and $\omega_{ij}^t$ denotes the edge weight reflecting long-term dynamics. Both $tp_{ij}^t$ and $\omega_{ij}^t$ are modeled based on the differences between adjacent time slice snapshots $\mathcal{G}_{t-1}$ and $\mathcal{G}_t$, thereby capturing the temporal dynamics of the graph structure. 
\end{definition}

Specifically, to implement this concept, we employ two key mechanisms.
Firstly, as depicted in the left section of Figure \ref{fig:framework}, multi-relational edge type modeling utilizes a Markov process to assign a unique type $tp_{ij}^t$ to each edge $e_{ij}^t \in \hat{E}_t$. The value of $tp$ is drawn from the set $\{\mathcal{e}, \mathcal{p}, \mathcal{d}\}$, where $\mathcal{e}$ denotes an emerging edge, $\mathcal{p}$ symbolizes a persisting edge, and $\mathcal{d}$ signifies an edge that has disappeared from the previous snapshot.
This process is mathematically expressed as follows:
\begin{gather}
    \hat{E}_t = E_{t-1} \cup E_{t} \label{equ:3} \\
    tp_{ij}^t = \begin{cases}
 \mathcal{e} & \text{if}\ e^t_{ij} \in E_{t} - E_{t-1}   \\
 \mathcal{p} & \text{if}\ e^t_{ij} \in E_{t} \cap E_{t-1} \\
 \mathcal{d} & \text{if}\ e^t_{ij} \in E_{t-1} - E_{t} 
 \end{cases} \label{equ:4}
\end{gather}
where Eq.\ref{equ:3} combines edges from consecutive time slices, while Eq.\ref{equ:4} categorizes these edges based on their temporal behavior. This categorization allows us to explicitly model the dynamic nature of edges, including those that have disappeared, which is crucial for understanding the evolving graph structure. In $\hat{\mathcal{G}}_t$, we provisionally reconstruct the edges that vanished between time slices $t-1$ and $t$. This intervention is based on the premise that the disappearance of an edge does not necessarily signal an immediate termination of inter-influence between the connected nodes. Conversely, the disappearance of an edge might potentially provoke a spectrum of impacts on the two nodes. Consequently, it is crucial for subsequent feature extraction models to explicitly recognize this aspect, as it significantly contributes to comprehending the dynamic nature of the graph.

Secondly, edge weight temporal updating captures the long-term dynamics by assigning and updating weights to edges based on their historical persistence and strength. 
Specifically, in examining the association between the duration of an edge and the correlation strength of nodes, the edge weight temporal updating procedure assigns a strength coefficient, denoted as $\omega$, to each edge. This coefficient is directly proportional to the duration of the edge, signifying that a longer edge duration corresponds to a higher strength coefficient:
\begin{equation}
    \omega_{ij}^t = \begin{cases}
  \alpha k^{\beta} & \text{ if } tp_{ij}^t= \mathcal{e} \ \text{or}\ \mathcal{p} \\
 \omega_{ij}^{t-1} & \text{ if } tp_{ij}^t = \mathcal{d} 
\end{cases}
\label{equ:weight}
\end{equation}
where $\omega^t_{ij}$ denotes the strength coefficient of the edge $e_{ij}^t$ at time slice $t$, and $k$ is indicative of the edge duration up to the point $t$. The parameters $\alpha$ and $\beta$ control the scaling and growth rate of edge weights respectively. These parameters can be tuned to adjust the model's sensitivity to edge duration, allowing for flexibility in different application scenarios.

As a result, the dynamic graph $\hat{\mathcal{G}_t}$ conserves the original topology while concurrently modeling the temporal state of the edge in a collaborative manner, taking into account the varying edge types and weights. This approach ensures a comprehensive representation of the edge dynamics across different time slices, uniquely capturing both the immediate changes (through edge types) and cumulative effects (through edge weights) of interactions.

\subsection{Recurrent Temporal Node Feature Extraction}

\begin{figure*}[tb]
    \centering
    \includegraphics[width=\textwidth]{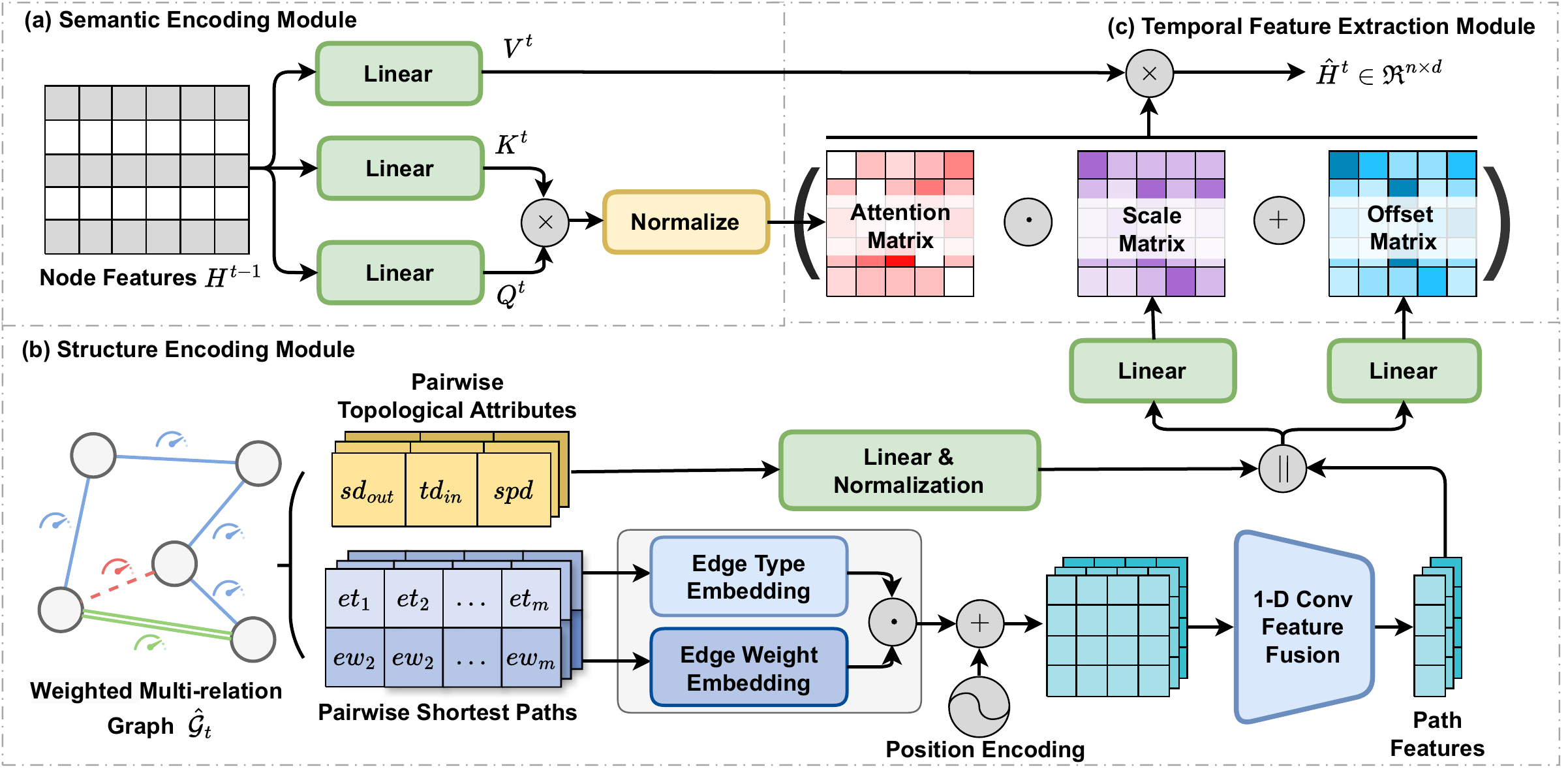}
    \caption{Structure-reinforced Graph Transformer for temporal node feature extraction. (a) Semantic Encoding Module: capturing global semantic correlations between nodes. (b) Structure Encoding Module: extracting topological dependencies from the weighted multi-relation difference graph, incorporating both original graph structure and temporal information. (c) Temporal Feature Extraction Module: fusing semantic and enhanced topological correlations to generate time-aware node representations.}
    \label{fig:model}
\end{figure*}

The Recurrent Temporal Node Feature Extraction process is the key component of our RSGT model, designed to capture both the temporal semantic and structural information of nodes in a dynamic graph setting. This process proposes a structure-reinforced graph transformer (SGT) to iteratively update node representations along the temporal dimension, integrating global semantic correlations, topological dependencies, and edge temporal states across consecutive snapshots.
Figure \ref{fig:framework} illustrates the process of recurrent temporal node feature extraction over the constructed weighted multi-relation difference graph $\hat{\mathcal{G}}_t$. For a sequence of graph snapshots $\{\hat{\mathcal{G}}_1, \hat{\mathcal{G}}_2, ..., \hat{\mathcal{G}}_T\}$, the SGT iteratively updates node representations as follows:
\begin{gather}
    \hat{\textbf{H}}^t = \mathcal{F}_{g_l}(\hat{\mathcal{G}}_t, \textbf{H}^{t-1}), \quad t = 1, 2, ..., T 
    \label{eq:recurrent} \\
    \textbf{H}^t = \hat{\textbf{H}}^t + \textbf{H}^{t-1} \label{eq:residual}
\end{gather}
where $\textbf{H}^t \in \Re^{|V|\times d}$ is the node feature matrix at time slice $t$, $\mathcal{F}_{g_l}(\cdot)$ denotes an $l$-layer SGT that constructed by stacking $l$ encoding layers, $|V|$ denotes the maximum node count across all time slices, and $d$ represents the dimensionality of the temporal node representation. Eq.\ref{eq:residual} implements a residual connection, facilitating gradient flow and enabling the model to leverage both current and historical information.

For the initial time slice ($t=1$), the SGT uses an initial embedding matrix $\textbf{H}^0=\textbf{X}\in \Re^{|V|\times d}$, where each row represents the initial feature vector of a particular node. This initial embedding can be derived from various sources, including node ID embeddings, where each node is assigned a unique ID embedded into a $d$-dimensional space; node attribute embeddings, which encode associated attributes such as user profiles in a social network; and pre-trained embeddings, which leverage pre-existing node representations from other models or domain knowledge as a starting point.

Next we will detail the core operations of our SGT, as shown in Figure \ref{fig:model}, including global semantic encoding, structure encoding, and temporal feature extraction.

\paragraph{\textbf{Global Semantic Encoding}}

The global semantic encoding module, illustrated in the upper left portion of Figure \ref{fig:model}, is designed to capture the overall semantic relationships between nodes across the entire graph, regardless of their topological distance. This global perspective allows the model to understand node similarities and interactions beyond immediate neighborhoods, which is particularly crucial in dynamic graphs where distant nodes may become relevant over time.

To achieve this global semantic understanding, we employ a standard self-attention mechanism \cite{vaswani2017attention} to compute semantic correlations between all node pairs. This approach allows each node to attend to every other node in the graph, effectively creating a fully connected semantic graph layered on top of the existing topological structure. 
The module takes $\textbf{H}^{t-1}$ as input and processes it as follows:
\begin{gather}
    \textbf{A}^t = \frac{\textbf{Q}^t(\textbf{K}^t)^T}{\sqrt{d_K}} = \frac{{\textbf{H}^{t-1}\textbf{W}_Q^t}({\textbf{H}^{t-1}\textbf{W}_K^t})^T}{\sqrt{d_K}} \label{eq:attention} \\
    \textbf{V}^t = \textbf{H}^{t-1}\textbf{W}_V^t \label{eq:value} 
\end{gather}
where $\textbf{W}_Q^t, \textbf{W}_K^t\in \Re^{d\times d'}$, and $\textbf{W}_V^t\in \Re^{d\times d}$ are trainable projection matrices. $\textbf{Q}^t$, $\textbf{K}^t$, and $\textbf{V}^t$ represent the query, key, and value matrices, respectively. The attention matrix $\textbf{A}^t\in \Re^{|V|\times |V|}$ captures the semantic correlation strength between nodes, where $a_{ij}^t \in \textbf{A}^t$ represents the correlation between nodes $v_i$ and $v_j$. For simplicity, we assume $d'=d$ and consider single-head self-attention, though extension to multi-head attention is straightforward.

\paragraph{\textbf{Structure Encoding}}
While global semantic encoding captures overall node relationships, the topological structure of the graph provides crucial information for node representation learning \cite{chen2022structure, ying2021transformers}. Our structure encoding module, depicted in the bottom right of Figure \ref{fig:model}, incorporates this topological information into the transformer architecture, leveraging both local and global structural features to enhance the model's capacity for capturing complex graph dynamics.

The judicious selection of topological features is paramount in effectively representing the structural characteristics of nodes within a graph. 
Drawing from seminal works in network analysis \cite{borgatti2005centrality, newman2018networks}, we incorporate node degree centrality and path-based measures, which have been demonstrated to encapsulate critical aspects of network topology. Specifically, node degree centrality, encompassing both in-degree and out-degree, serves as a fundamental metric reflecting a node's connectivity and potential influence within the network, while the shortest path length between nodes captures structural proximity and information flow efficiency, thereby providing a comprehensive representation of a node's position and role within the graph structure.
Thus, for each node pair $\langle v_i, v_j \rangle$, we construct a topological attribute vector $\textbf{attr}_s$:
\begin{equation}
    \textbf{attr}_{(ij)s}^t = [sd_{out}, td_{in}, spd] \label{eq:topo_attr}
\end{equation}
where $sd_{out}$, $td_{in}$, and $spd$ denote the out-degree of the source node, the in-degree of the target node, and the shortest path length between the nodes, respectively.
Then we employ a transformation mechanism that projects this attribute vector into a higher-dimensional space, followed by a normalization step to ensure stable training dynamics:
\begin{equation}
    \textbf{r}^t_{(ij)s} = \text{LayerNorm}(\textbf{W}_s \textbf{attr}_{(ij)s}^t + \textbf{b}_s) \label{eq:topo_transform}
\end{equation}
where $\textbf{W}_s \in \Re^{d_e \times 3}$ and $\textbf{b}_s \in \Re^{d_e}$ are learnable parameters, and $\text{LayerNorm}$ denotes layer normalization. The resulting representation $\textbf{r}^t_{(ij)s} \in \Re^{d_e}$ encapsulates the complex topological relationships between nodes, facilitating the model's ability to discern intricate structural patterns within the dynamic graph.

To additionally represent edge temporal states, we establish a shortest path matrix $\textbf{ATTR}_{(ij)p}^t \in \Re^{2\times spd}$ for each node pair, since it not only encodes the topological relationship between nodes but also incorporates information about how edge types and weights change over time, which is essential for understanding the dynamic nature of the graph:
\begin{gather}
    \textbf{ATTR}_{(ij)p}^t = [\textbf{attr}_e^t; \textbf{attr}_w^t] \label{eq:path_matrix}
\end{gather}
where $\textbf{attr}_e^t$ represents the edge types along the path $\mathcal{p}_{v_i \rightarrow v_j}^t$ at time $t$, and $\textbf{attr}_w^t$ represents the corresponding edge weights.
To transform these discrete attributes into continuous representations suitable for neural network processing, we employ embedding layers:
\begin{gather}
    \textbf{attr}_e^t \mapsto \textbf{R}_e^t \in \Re^{spd\times d_e} \label{eq:edge_embed}\\
    \textbf{attr}_w^t \mapsto \textbf{R}_w^t \in \Re^{spd\times d_e} \label{eq:weight_embed}
\end{gather}
where $d_e$ is the embedding dimension. These embeddings allow us to learn rich, continuous representations of edge types and weights, facilitating more nuanced modeling of temporal dynamics. 

To capture the varying intensities of different edge types in the graph's temporal evolution, we integrate edge type features with their corresponding weights. This integration is achieved through the Hadamard product, which effectively modulates the impact of each edge type based on its strength:
\begin{gather}
    \textbf{R}_p^t = \textbf{R}_e^t \odot \textbf{R}_w^t \label{eq:path_feature}
\end{gather}
where $\odot$ denotes the Hadamard product.

Since the position of edges along a path in a graph structure can carry crucial semantic information, and to enable our model to differentiate between identical edge types occurring at different positions, we incorporate positional encoding following the conventional Transformer design \cite{vaswani2017attention} that allows the model to consider the relative positions of edges, enhancing its ability to capture complex structural patterns:
\begin{gather}
    \textbf{PE}_{pos, 2i}=sin(\frac{pos}{10000^{2i/d_{e}}}) \label{eq:pe_sin}\\
    \textbf{PE}_{pos, 2i+1}=cos(\frac{pos}{10000^{2i/d_{e}}}) \label{eq:pe_cos}\\
    \hat{\textbf{R}}_p^t = \textbf{R}_p^t + \textbf{PE} \label{eq:pe_add}
\end{gather}
where $pos$ is the position of an edge in the path and $i$ is the column index.

We then apply a one-dimensional convolutional layer to reduce the dimensionality of $\textbf{R}_p^t$, resulting in the final path feature $\hat{\textbf{r}}_{(ij)p}^t \in \Re^{d_e}$. The graph structure feature $\textbf{r}_{ij}^t\in \Re^{2d_e}$ is formed by concatenating the topology feature and the path feature:
\begin{gather}
    \textbf{r}_{ij}^t = \textbf{r}^t_{(ij)s} \oplus \hat{\textbf{r}}_{(ij)p}^t \label{eq:graph_structure}
\end{gather}
where $\oplus$ denotes vector concatenation.

\paragraph{\textbf{Temporal Feature Extraction}}

To incorporate the structural information into the transformer model, we propose a structure-aware attention mechanism, as shown in the upper right section of Figure \ref{fig:model}. This mechanism modulates the standard attention score $a_{ij}^t\in \textbf{A}^t$ based on the graph's structural features $\textbf{r}_{ij}^t$:
\begin{gather}
    \lambda_{ij}^t = \textbf{W}_s(\textbf{r}^t_{ij})^T \label{eq:scale}\\
    \sigma_{ij}^t = \textbf{W}_{\sigma}(\textbf{r}^t_{ij})^T \label{eq:offset}\\
    \hat{a}_{ij}^t = \lambda_{ij}^ta_{ij}^t+\sigma_{ij}^t \label{eq:modulated_attention}\\
    \hat{\textbf{H}}^t = Normalize(\hat{\textbf{A}}^t)\textbf{V} \label{eq:final_output}
\end{gather}
where $\lambda_{ij}^t$ and $\sigma_{ij}^t$ are scale and offset coefficients, respectively. $\hat{a}_{ij} \in \hat{\textbf{A}}^t$ is the structure-reinforced attention for nodes $v_i$ and $v_j$. Thus the final output, i.e. the temporal feature $\hat{\textbf{H}}^t \in \Re^{|V|\times d}$, encodes global semantic correlation, topological dependency, and edge temporal states.

\subsection{Model Training}

To facilitate the training and optimization of model parameters, we incorporate dynamic link prediction as an auxiliary task. For a candidate edge $e_{ij}^{t+1}$, we compute its feature $\textbf{h}_{e_{ij}}^{t+1} \in \Re^{d}$ as the absolute difference of node features: $|\textbf{h}_i^t-\textbf{h}_j^t|$, which captures the relative differences in node representations and is crucial for predicting potential connections. The edge feature is then input into a fully-connected neural network layer for binary classification:
\begin{equation}
    \hat{p}_{ij}^{t+1} = \sigma(\textbf{W}_o\textbf{h}_{e_{ij}}^{t+1}+\textbf{b}_o) 
\end{equation}
where $\textbf{W}_o$ and $\textbf{b}_o$ are trainable weights, and $\hat{p}_{ij}^{t+1}$ denotes the predicted probability of the edge being positive.
We then adopt cross entropy as the loss function:
\begin{gather}
    J_{ij}=-p_{ij}^{t+1}log\hat{p}_{ij}^{t+1}-(1-p_{ij}^{t+1})log(1-\hat{p}_{ij}^{t+1})\\
    J = \mathbb{E}(\sum_{i,j\in V}J_{ij}) + \lambda ||\Theta||^2_2
\end{gather}
where $\Theta$ encompasses all trainable parameters, and $\lambda$ is a hyperparameter controlling the $L2$ regularization strength to mitigate overfitting.

For parameter optimization, we employ the mini-batch AdamW optimizer \cite{loshchilov2017decoupled}, which, by virtue of its efficient handling of large-scale graph data and its capacity to adapt learning rates dynamically, proves particularly well-suited to our model, where diverse parameter types and the evolving nature of graphs necessitate an approach that can accommodate both structural complexity and temporal variability.

\subsection{Complexity Analysis}

The time complexity of RSGT can be analyzed by examining its three main components. First, the Edge Temporal State Modeling, which constructs the weighted multi-relation graph, has a complexity of $O(|E|)$, where $|E|$ is the number of edges. This includes both edge type assignment and weight updating processes. The second and core component, Recurrent Temporal Node Feature Extraction, involves more intricate computations. For a graph with $|V|$ nodes and $d$-dimensional features, the global semantic encoding requires $O(|V|^2d)$ operations for self-attention computation. The structure encoding process, which includes subgraph sampling, shortest path computation, topological attribute calculation, and feature transformation, has a complexity of $O(|V|k(\log k + d))$, where $k$ is the subgraph sampling size. Notably, the use of subgraph sampling significantly reduces the computational burden from $O(|V|^2\log|V|)$ to $O(|V|k\log k)$, as typically $k \ll |V|$. The structure-aware attention mechanism adds an additional $O(|V|^2d)$ complexity. Considering $l$ layers in the SGT, the complexity for a single time step becomes $O(l(|V|k(\log k + d) + |V|^2d))$. Finally, the model training for dynamic link prediction contributes $O(|E|d)$ to the overall complexity. Combining these components and considering $T$ time steps, the total time complexity of RSGT is
\begin{equation}
    O(Tl(|V|k(log k + d) + |V|^2d) + |E|d)
\end{equation}

It's worth noting that while the worst-case complexity appears to be quadratic in the number of nodes due to the global semantic encoding, our use of subgraph sampling significantly reduces the computational burden in practice, especially for the structure encoding part. This optimization allows RSGT to efficiently process large-scale dynamic graphs. In real-world scenarios where typically $k \ll |V|$ and $d \ll |V|$, the effective complexity is often much lower than the theoretical upper bound, enabling RSGT to scale well with increasing graph sizes.

%% file: experiments.tex
\section{Experiments}
\label{sec:exp}
In order to substantiate the efficacy of the proposed RSGT model, we conduct a series of rigorous experimental assessments. These experiments are performed using a workstation furnished with two NVIDIA GTX 1080Ti GPUs, an Intel(R) Xeon(R) Gold 6130 processor clocked at 2.60 GHz, and 192GB of RAM. The RSGT model's components are implemented using Python 3.7.1 and Pytorch 1.4.0.

\subsection{Datasets}
\begin{table}[tb]
\centering
\caption{Statistics of the four datasets.}
\label{tab:datasets}
\begin{tabular}{@{}cccc@{}}
\toprule
Dataset & \# of Nodes & \# of Edges & Train/Test splits\\ \midrule
twi-Tennis & 1,000 & 40,839 & 100/20 \\
CollegeMsg & 1,899 & 59,835 & 25/63\\
cit-HepTh & 7,577 & 51,315 & 77/1\\
sx-MathOF & 24,818 & 506,550 & 64/15\\
\bottomrule
\end{tabular}
\end{table}

To evaluate RSGT's performance across diverse scenarios, we conducted experiments on four real-world dynamic graph datasets: twi-Tennis, CollegeMsg, cit-HepTh, and sx-MathOF. These datasets span various domains and exhibit distinct characteristics in terms of time scales, node and edge densities, and interaction types.

The twi-Tennis dataset \cite{beres2018temporal} captures Twitter mentions during major tennis tournaments, with nodes representing accounts and edges denoting mentions. CollegeMsg \cite{2009patterns} is derived from a university's online social network, where edges represent private messages between users. The cit-HepTh dataset \cite{2005grpahsovertime} is based on arXiv's high energy physics papers, with nodes as papers and edges as citations. Lastly, sx-MathOF \cite{paranjape2017motifs} represents interactions on the Math Overflow forum, with edges indicating responses or comments to questions.

Each dataset is divided into multiple snapshots, capturing the temporal evolution of the respective networks. Node features, where available, are represented through appropriate embeddings. Table \ref{tab:datasets} summarizes the key statistics of these datasets.
\begin{table*}[t]
\centering
\caption{Performance comparisons of dynamic link prediction among competing baselines on twi-Tennis and ColledgeMsg.}
\label{tab:result}
\begin{tabularx}{\textwidth}{XXXXXXXXXX}
\toprule
     & \multicolumn{4}{c}{twi-Tennis} & \multicolumn{4}{c}{CollegeMsg} \\ \cmidrule(l){2-9} 
                     & Accuracy $\uparrow$          & Recall $\uparrow$         & Precision $\uparrow$ & F1 $\uparrow$ & Accuracy $\uparrow$ & Recall $\uparrow$ & Precision $\uparrow$ & F1 $\uparrow$ \\ \midrule
DeepWalk & 61.96±1.27 & 61.05±2.83 & 61.51±2.62 & 61.27±2.81 & 66.54±5.36 & 67.57±5.81 & 68.22±5.95 & 67.86±5.86 \\
GraphSAGE & 62.76±1.76 & 62.02±1.63 & 62.50±1.78 & 62.26±1.66 & 58.91±3.67 & 60.23±4.15 & 60.57±4.30 & 60.45±4.22 \\ 
EvolveGCN & 64.73±0.64 & 63.51±0.92 & 64.12±1.01 & 63.80±0.98 & 63.27±4.42 & 65.62±4.64 & 65.37±4.81 & 65.44±4.72 \\
CoEvoSAGE & 67.39±2.51 & 67.71±2.03 & 68.24±2.17 & 67.95±2.12 & 67.72±3.17 & 69.81±3.12 & 70.20±3.25 & 70.02±3.64 \\
ROLAND  & 68.45±2.08 & 68.61±1.74 & 69.03±1.81 & 68.80±1.74 & 69.44±2.86 & 70.19±2.58 & 70.53±2.63 & 70.32±2.64 \\ 
CTDNE & 58.14±2.67 & 58.33±2.04 & 58.91±2.19 & 58.61±2.12 & 62.55±3.67 & 65.34±2.20 & 65.82±2.36 & 65.56±2.34 \\
TGAT & 69.01±1.54 & 69.03±1.41 & 69.56±1.56 & 69.24±1.47 & 70.35±2.54 & 71.01±2.35 & 71.43±2.46 & 71.20±2.37 \\
CAW & 71.35±1.68 & 71.15±1.41 & 72.03±1.46 & 71.57±1.42 & 73.04±2.11 & 72.71±1.95 & 73.23±2.01 & 72.95±1.98 \\
TREND & 74.02±1.78 & 73.83±1.61 & 75.46±1.63 & 74.63±1.66 & 74.55±1.95 & 74.33±1.91 & 75.91±2.04 & 75.64±2.09 \\
DyGFormer & \cellcolor[HTML]{C0C0C0}75.89±1.52 & \cellcolor[HTML]{C0C0C0}75.67±1.43 & \cellcolor[HTML]{C0C0C0}77.21±1.58 & \cellcolor[HTML]{C0C0C0}76.43±1.50 & \cellcolor[HTML]{C0C0C0}76.32±1.78 & \cellcolor[HTML]{C0C0C0}76.11±1.69 & \cellcolor[HTML]{C0C0C0}77.54±1.86 & \cellcolor[HTML]{C0C0C0}76.82±1.77 \\ 
R-GSAGE & 74.85±1.63 & 74.62±1.55 & 76.18±1.59 & 75.39±1.57 & 75.21±1.83 & 74.98±1.75 & 76.43±1.92 & 75.70±1.83 \\ 
ConTIG & 75.13±1.59 & 74.91±1.48 & 76.49±1.55 & 75.69±1.51 & 75.78±1.81 & 75.54±1.72 & 77.02±1.89 & 76.27±1.80 \\
\midrule
\textbf{RSGT} & \textbf{87.59±0.55} & \textbf{87.13±0.52} & \textbf{88.04±0.67} & \textbf{87.55±0.55} & \textbf{86.81±0.14} & \textbf{86.36±0.19} & \textbf{87.21±0.21} & \textbf{86.70±0.13} \\
\textit{Gains} & 15.42\% & 15.14\% & 14.03\% & 14.55\% & 13.74\% & 13.47\% & 12.47\% & 12.86\% \\ \bottomrule
\end{tabularx}
\end{table*}
\begin{table*}[h!]
\centering
\caption{Performance comparisons of dynamic link prediction among competing baselines on cit-HepTh and sx-MathOF.}
\label{tab:result2}
\begin{tabularx}{\textwidth}{XXXXXXXXXX}
\toprule
     & \multicolumn{4}{c}{cit-HepTh} & \multicolumn{4}{c}{sx-MathOF} \\ \cmidrule(l){2-9} 
                     & Accuracy $\uparrow$          & Recall $\uparrow$         & Precision $\uparrow$ & F1 $\uparrow$ & Accuracy $\uparrow$ & Recall $\uparrow$ & Precision $\uparrow$ & F1 $\uparrow$ \\ \midrule
DeepWalk & 51.55±0.90 & 49.89±0.92 & 50.89±0.89 & 50.39±0.98 & 66.23±0.89 & 66.47±0.94 & 67.81±0.98 & 67.14±1.12 \\
GraphSAGE & 70.72±1.96 & 70.56±2.12 & 71.98±2.04 & 71.27±2.41 & 65.32±1.55 & 65.52±1.35 & 66.84±1.69 & 66.18±1.21 \\ 
EvolveGCN & 61.57±1.53 & 61.80±1.52 & 63.04±1.44 & 62.42±1.54 & 68.35±0.68 & 68.33±0.58 & 69.71±0.39 & 69.02±0.35 \\
CoEvoSAGE & 65.88±1.32 & 65.98±1.28 & 67.32±1.19 & 66.65±0.98 & 71.38±1.02 & 71.37±0.92 & 72.81±0.87 & 72.09±0.96 \\
ROLAND  & 70.57±1.51 & 70.61±1.44 & 72.03±1.28 & 71.32±1.52 & 73.35±0.98 & 72.31±0.82 & 73.77±0.69 & 73.04±0.85 \\ 
CTDNE & 49.42±1.86 & 43.79±1.98 & 44.67±2.67 & 44.23±2.92 & 60.72±0.28 & 60.30±0.39 & 61.52±0.19 & 60.91±0.51 \\
TGAT & 71.42±1.28 & 70.39±1.30 & 71.81±1.36 & 71.10±1.34 & 74.15±1.01 & 74.00±0.81 & 75.50±0.83 & 74.75±0.79 \\
CAW & 72.75±1.19 & 71.78±1.21 & 73.22±1.28 & 72.50±1.23 & 77.95±0.77 & 76.89±0.76 & 78.45±0.65 & 77.67±0.72 \\
TREND & 80.37±2.08 & 80.32±1.97 & 81.94±1.82 & 81.13±1.92 & 79.82±1.56 & 79.21±1.27 & 80.81±1.31 & 80.01±1.34 \\ 
DyGFormer & 81.25±1.89 & 81.18±1.82 & 82.73±1.75 & 81.95±1.78 & 80.94±1.43 & 80.36±1.21 & 81.89±1.28 & 81.12±1.24 \\ 
R-GSAGE & 81.58±1.85 & 81.52±1.79 & 83.08±1.71 & 82.29±1.75 & \cellcolor[HTML]{C0C0C0}81.63±1.38 & \cellcolor[HTML]{C0C0C0}81.02±1.17 & \cellcolor[HTML]{C0C0C0}82.57±1.23 & \cellcolor[HTML]{C0C0C0}81.79±1.20 \\ 
ConTIG & \cellcolor[HTML]{C0C0C0}82.19±1.92 & \cellcolor[HTML]{C0C0C0}82.13±1.85 & \cellcolor[HTML]{C0C0C0}83.72±1.78 & \cellcolor[HTML]{C0C0C0}82.92±1.81 & 81.21±1.41 & 80.62±1.19 & 82.16±1.25 & 81.38±1.22 \\
\midrule
\textbf{RSGT} & \textbf{87.20±0.49} & \textbf{86.85±0.46} & \textbf{87.49±0.32} & \textbf{87.17±0.40} & \textbf{87.91±0.32} & \textbf{87.45±0.38} & \textbf{88.35±0.46} & \textbf{87.90±0.42} \\
\textit{Gains} & 6.10\% & 5.75\% & 4.50\% & 5.13\% & 7.69\% & 7.94\% & 7.00\% & 7.47\% \\\bottomrule
\end{tabularx}
\end{table*}

\subsection{Competing Methods}
To assess the efficacy of our proposed RSGT, we compare it with twelve state-of-the-art benchmarks, encompassing both static and dynamic graph representation learning approaches:
\begin{itemize}
    \item \textbf{DeepWalk} \cite{perozzi2014deepwalk}: It treats random walk sequences as sentences and employs skip-grams \cite{mikolov2013distributed} to learn node representations.
    \item \textbf{GraphSAGE} \cite{hamilton2017inductive}: It is a generalized inductive framework built on GNN. 
    \item \textbf{EvolveGCN} \cite{pareja2020evolvegcn}: It uses RNN to evolve GCN parameters to encapsulate the dynamic information of discrete dynamic graphs.
    \item \textbf{CoEvoSAGE} \cite{wang2021modeling}: It employs GraphSAGE to identify the structural dependencies and a temporal self-attention architecture to model long-range evolutionaries.
    \item \textbf{ROLAND} \cite{you2022roland}: It researchers in easily repurposing any static GNN to dynamic graphs. It views the node embeddings at different GNN layers as hierarchical node states and updates them recurrently over time.
    \item \textbf{CTDNE} \cite{nguyen2018continuous}: It generalizes random walk-based embedding techniques to continuous dynamic graphs.
    \item \textbf{TGAT} \cite{DBLP:conf/iclr/XuRKKA20}: It employs a temporal graph attention layer, leveraging self-attention and a novel time encoding to capture temporal-topological interactions. 
    \item \textbf{CAW} \cite{DBLP:conf/iclr/WangCLL021}: It proposes Causal Anonymous Walks to encapsulate temporal network dynamics without manually counting network motifs.
    \item \textbf{TREND} \cite{wen2022trend}: It models the Hawkes processes \cite{hawkes1971spectra} to incorporate the evolutionary properties of temporal edges into node representations from continuous dynamic graphs.
    \item \textbf{DyGFormer} \cite{yu2023towards}: It employs a Transformer-based architecture with neighbor co-occurrence encoding and patching techniques to learn from nodes' historical first-hop interactions.
    \item \textbf{R-GSAGE} \cite{yao2024recurrent}: It captures the continuous dynamic evolution of node embedding trajectories using ordinary differential equations and self-attention mechanisms.
    \item \textbf{ConTIG} \cite{wang2024contig}: It integrates a recurrent structure into GraphSAGE to jointly explore structural and temporal patterns while maintaining a lightweight architecture.
\end{itemize}

\subsection{Experiment Results and Analyses}
\subsubsection{Prediction task and parameter settings}
Following the experimental framework outlined in \cite{wen2022trend}, we use dynamic link prediction \cite{DBLP:conf/icml/SarkarCJ12} to evaluate the efficacy of the proposed RSGT model. 

Given a dynamic graph $\mathcal{G}$, we divide the snapshots from various time slices into training and test datasets. 
The training dataset $\mathcal{D}^{tr}$ encompasses all snapshots captured prior to time $tr$. These snapshots are utilized for training the representation learning models. Conversely, the snapshots captured post time $tr$ are allocated to the test dataset $\mathcal{D}^{te}$.
To comprehensively evaluate our model's capability in learning both short-term and long-term historical information, as well as its performance in predicting links over varying time horizons, we employ different train-test split ratios for each dataset, as shown in Tabel \ref{tab:datasets}. This diverse set of split ratios allows us to simulate various real-world scenarios, from rapidly evolving social networks to more stable citation networks, thereby providing a more thorough assessment of RSGT's adaptability and predictive power across different temporal scales.

Taking into consideration both the size and complexity of the datasets, we designate the dimension of node representation $d$ as 32 for the twi-Tennis, CollegeMsg, and sx-MathOF datasets, and 16 for the cit-HepTh dataset.
During the testing phase, we initially generate temporal node representations using the trained model. These representations are subsequently input into a logistic regression classifier to predict the likelihood of a candidate edge being either positive or negative, assessed one snapshot at a time. 
For each test snapshot in $\mathcal{D}^{te}$, we adopt a 1:1 ratio of negative sampling to construct the complete test edge set. Here, 80\% of the total test edges are utilized to train the classifier, while the remaining 20\% are reserved for testing purposes. 
The scale factor $\alpha$ and the power factor $\beta$ are kept constant at 1 in all datasets. 
We conduct the experiment five times and report the mean classification accuracy and F1 scores, along with their respective standard deviations.

\subsubsection{Comparison with competing baselines}
The experimental results for dynamic link prediction, as outlined in Table \ref{tab:result} and Table \ref{tab:result2}, reveal a comprehensive comparison amongst the various baseline methods. The sub-optimal performance is indicated by darker shading, while the most striking results are highlighted in bold. 
Our proposed method, RSGT, demonstrates consistently superior performance across all four experimental datasets when compared to both traditional and state-of-the-art techniques. This superiority is evident from the relative increments in Accuracy, which range from 6.10\% to 15.42\%, and F1 scores, showing improvements from 5.13\% to 14.55\%.

In comparing static and dynamic approaches, we observe that dynamic graph methods generally outperform those designed for static graphs, underscoring the importance of integrating temporal information for comprehending graph evolution. For instance, on the twi-Tennis dataset, the dynamic method CAW achieves an accuracy of 71.35\% and an F1 score of 71.57\%, significantly outperforming the static method GraphSAGE, which only reaches an accuracy of 62.76\% and an F1 score of 62.26\%. This trend is also evident in the CollegeMsg dataset, where CAW (accuracy: 73.04\%, F1: 72.95\%) substantially surpasses GraphSAGE (accuracy: 58.91\%, F1: 60.45\%). However, this pattern is not universal across all datasets. On the cit-HepTh dataset, the static method GraphSAGE (accuracy: 70.72\%, F1: 71.27\%) outperforms some dynamic approaches such as EvolveGCN (accuracy: 61.57\%, F1: 62.42\%) and CTDNE (accuracy: 49.42\%, F1: 44.23\%). This variability highlights the complexity of dynamic graph learning and the need for methods that can adapt to different types of temporal evolution.

The performance of different models also varies significantly across datasets, reflecting the impact of dataset characteristics on model efficacy. Social network datasets (twi-Tennis and CollegeMsg) favor models that excel in capturing short-term dependencies and frequent interactions. For example, on the twi-Tennis dataset, recent dynamic graph methods like DyGFormer show strong performance (accuracy: 75.89\%, F1: 76.43\%), likely due to their ability to model rapid changes in network structure. In contrast, the citation network (cit-HepTh) presents unique challenges due to its long-term dependencies and relatively slower evolution. Here, methods that can effectively capture long-range temporal dependencies, such as ConTIG, perform well (accuracy: 82.19\%, F1: 82.92\%). The Q\&A platform dataset (sx-MathOF), with its complex interaction patterns, favors models that can balance both structural and temporal aspects of the graph. On this dataset, R-GraphSAGE demonstrates strong performance (accuracy: 81.63\%, F1: 81.79\%), underscoring the importance of versatile approaches that can adapt to diverse interaction patterns.

Despite the varying performance of different methods across datasets, RSGT consistently outperforms all baselines. On the twi-Tennis dataset, RSGT achieves an accuracy of 87.59\% and an F1 score of 87.55\%, surpassing the best baseline (DyGFormer) by 15.42\% in accuracy and 14.55\% in F1 score. Similar performance gains are observed across all datasets, with RSGT showing improvements of 13.74\% in accuracy and 12.86\% in F1 score on CollegeMsg, 6.10\% in accuracy and 5.13\% in F1 score on cit-HepTh, and 7.69\% in accuracy and 7.47\% in F1 score on sx-MathOF, compared to the best performing baselines on each dataset.

RSGT's superior performance across all datasets can be attributed to several key innovations. Its edge temporal state modeling through a weighted multi-relation dynamic graph provides crucial heuristic information for node representation learning, while its adaptive learning mechanism mitigates the influence of erratic neighbors during recurrent updates. The structure-reinforced graph transformer architecture captures both global semantic correlations and local topological features, leading to high-quality node representations. Moreover, RSGT's comprehensive temporal modeling, combining recurrent structure with edge temporal state modeling, effectively captures both short-term and long-term dependencies. These features collectively contribute to RSGT's robustness, scalability, and superior inductive learning capability, enabling it to consistently outperform specialized methods across diverse dynamic graph scenarios. The model's ability to adapt to different graph evolution patterns, coupled with its effective modeling of both structural and temporal aspects, establishes RSGT as a powerful and versatile approach for dynamic graph representation learning.

\begin{figure*}[!t]
\centering
\subfloat[Accuracy]{\includegraphics[width=0.24\textwidth]{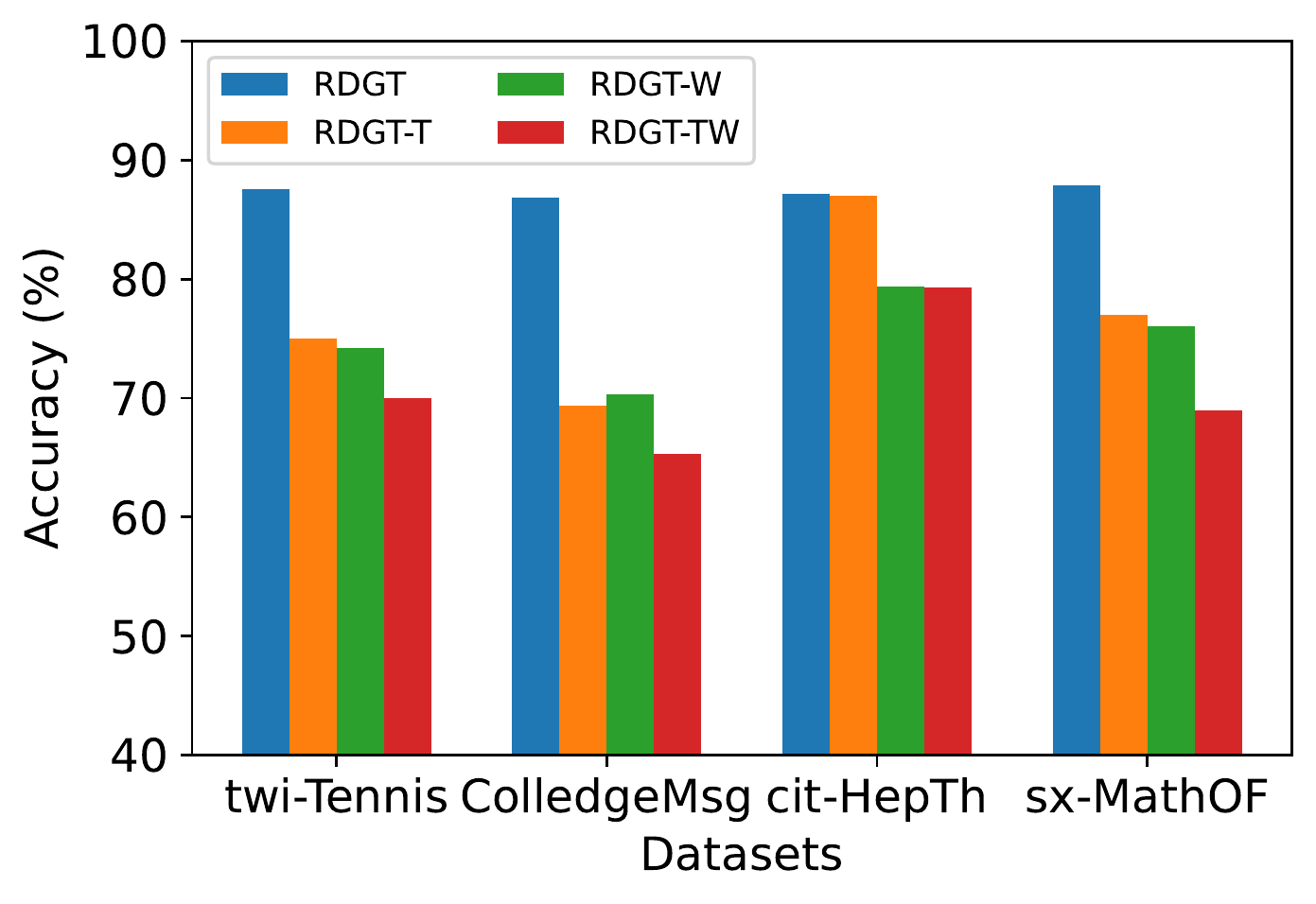}%
\label{fig:ac}}
\hfil
\subfloat[Recall]{\includegraphics[width=0.24\textwidth]{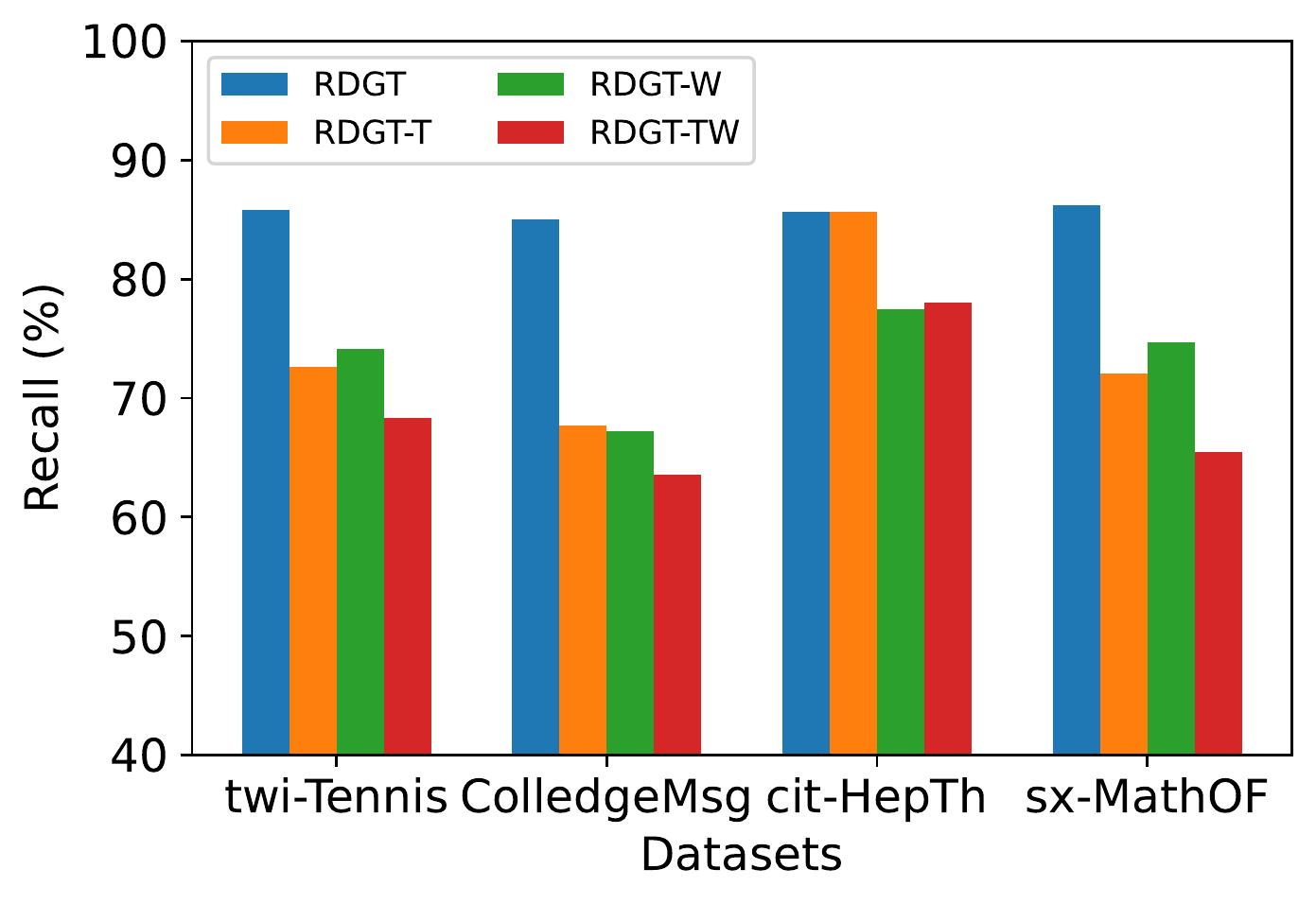}%
\label{fig:ar}}
\hfil
\subfloat[Precision]{\includegraphics[width=0.24\textwidth]{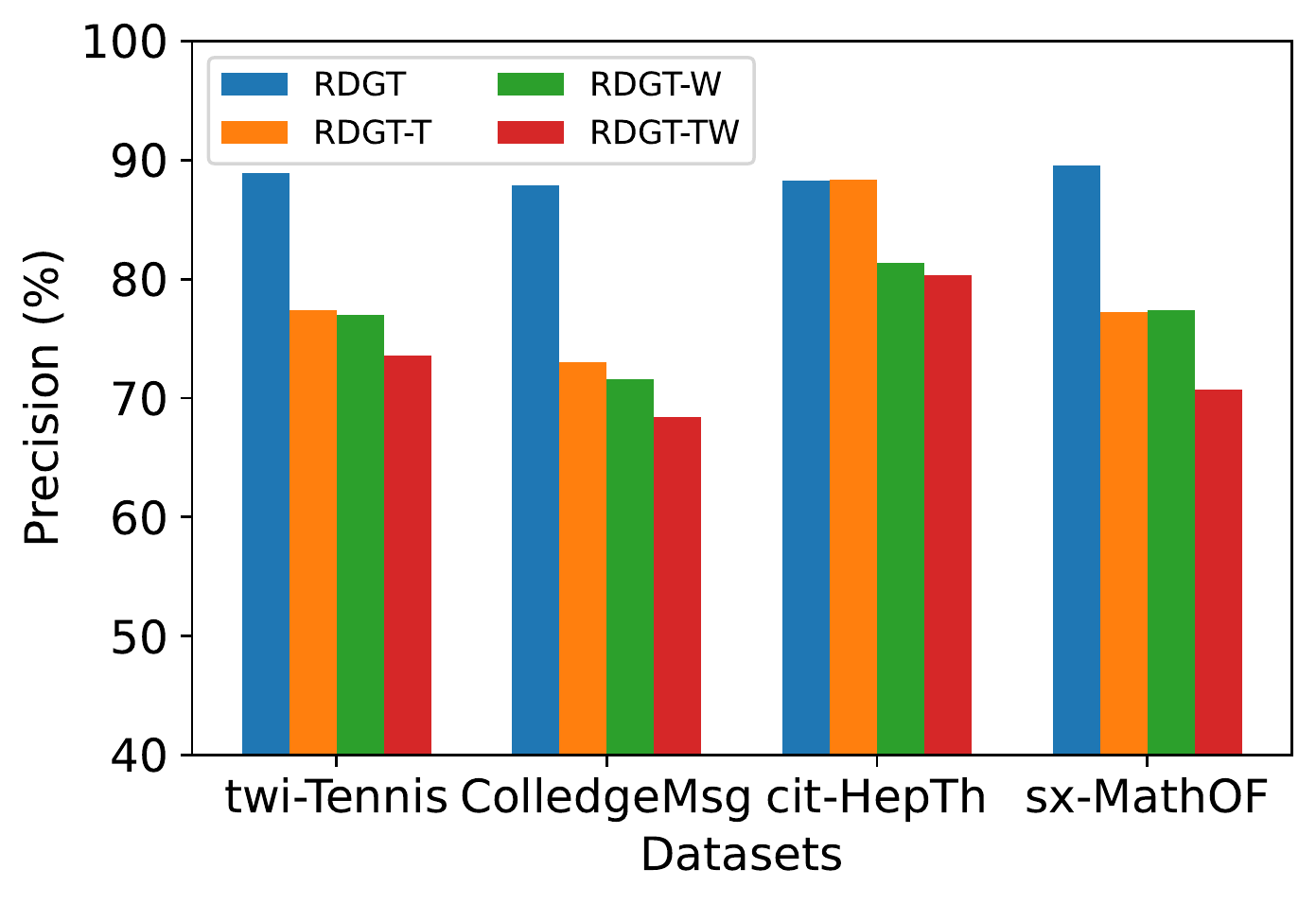}%
\label{fig:ap}}
\hfil
\subfloat[F1]{\includegraphics[width=0.24\textwidth]{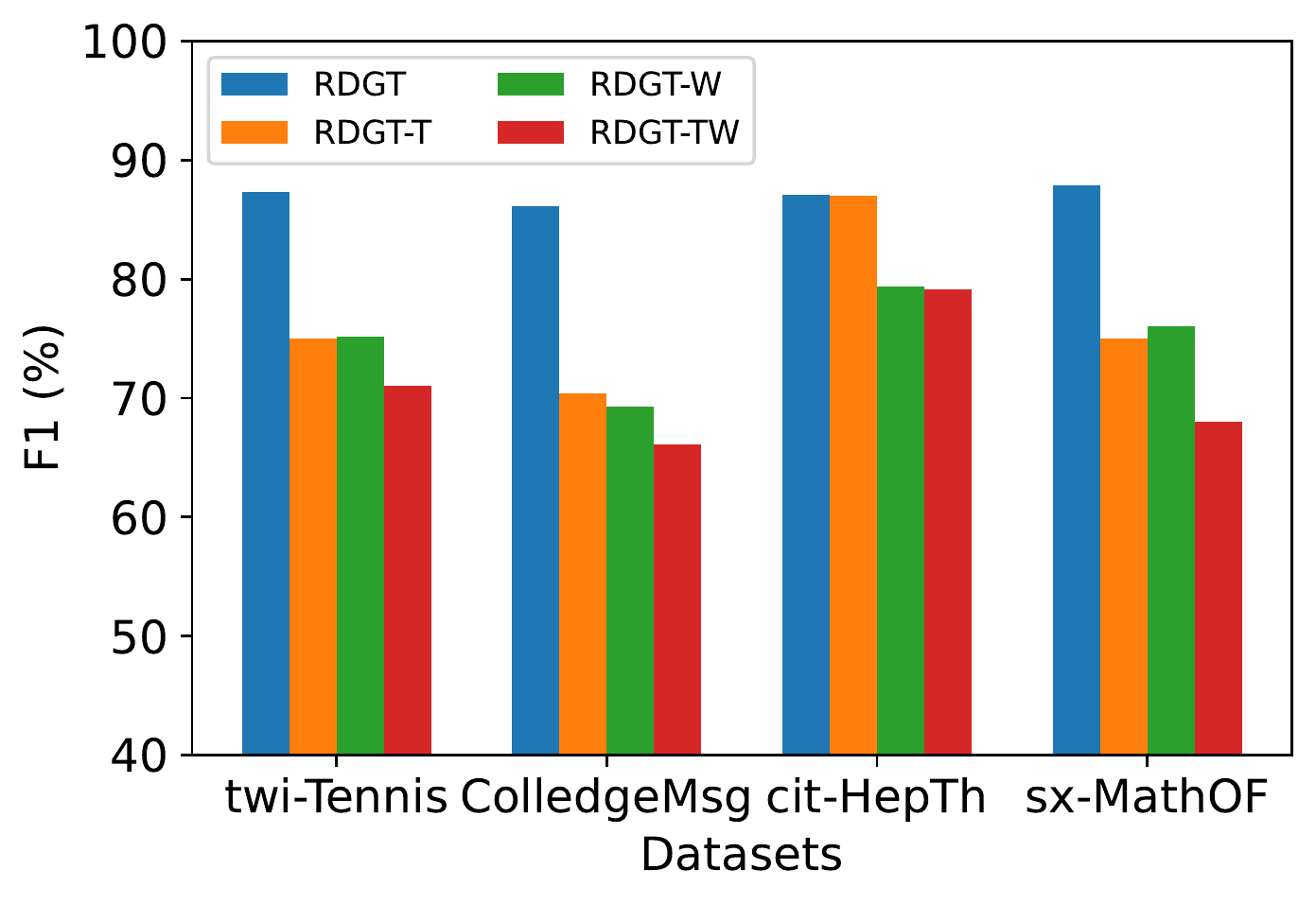}%
\label{fig:af}}
\hfil
\subfloat[Accuracy]{\includegraphics[width=0.24\textwidth]{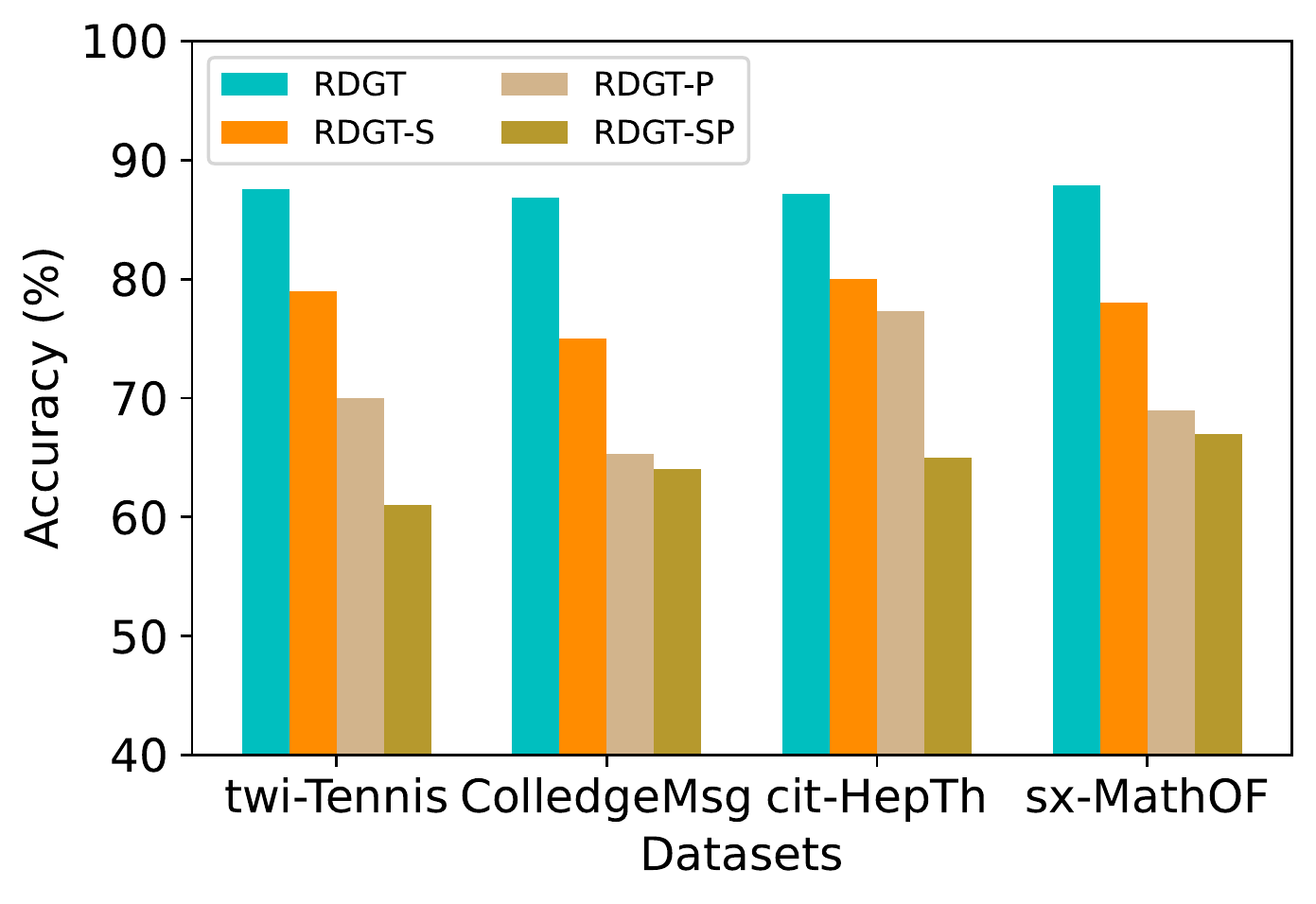}%
\label{fig:asc}}
\hfil
\subfloat[Recall]{\includegraphics[width=0.24\textwidth]{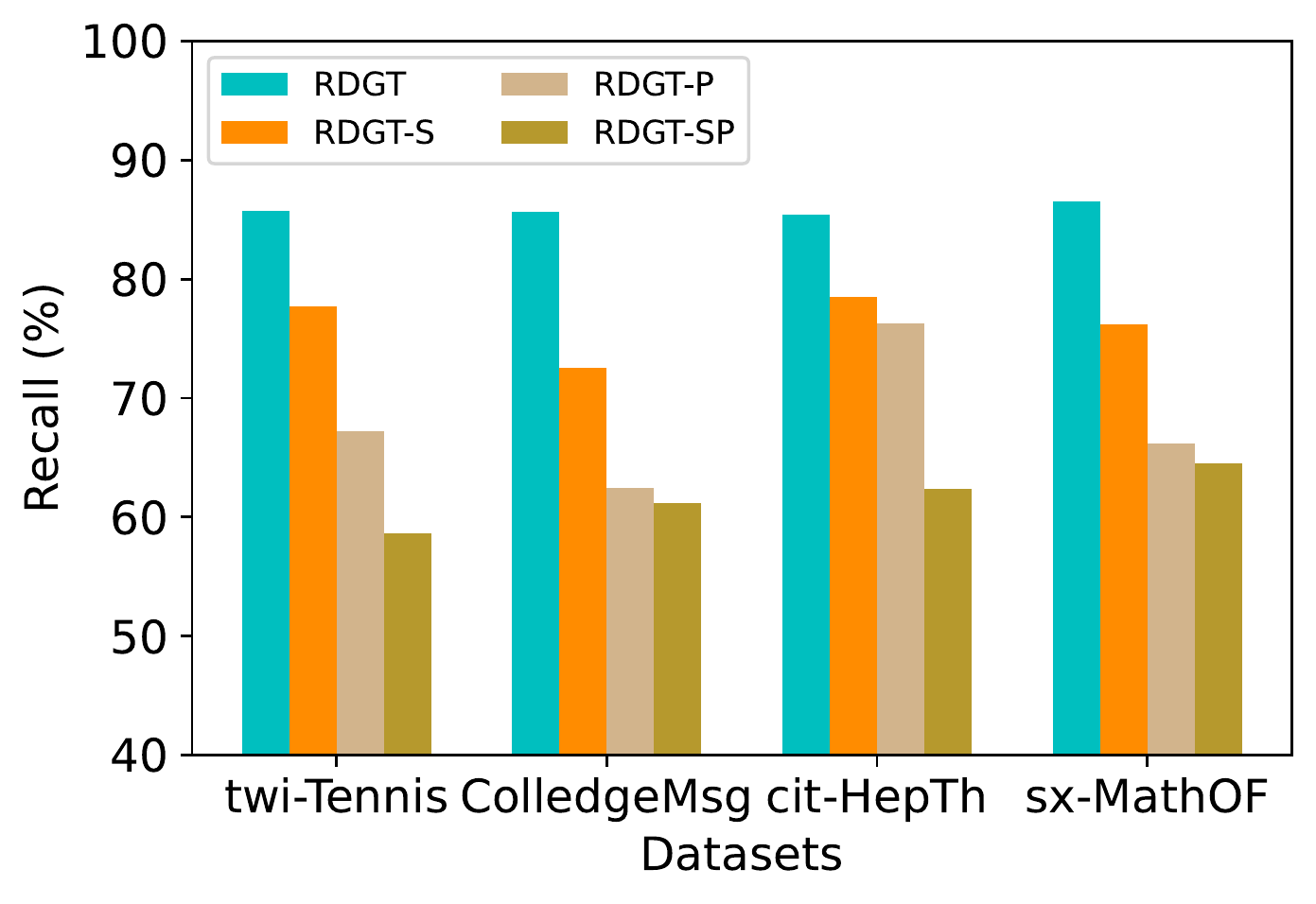}%
\label{fig:asr}}
\subfloat[Precision]{\includegraphics[width=0.24\textwidth]{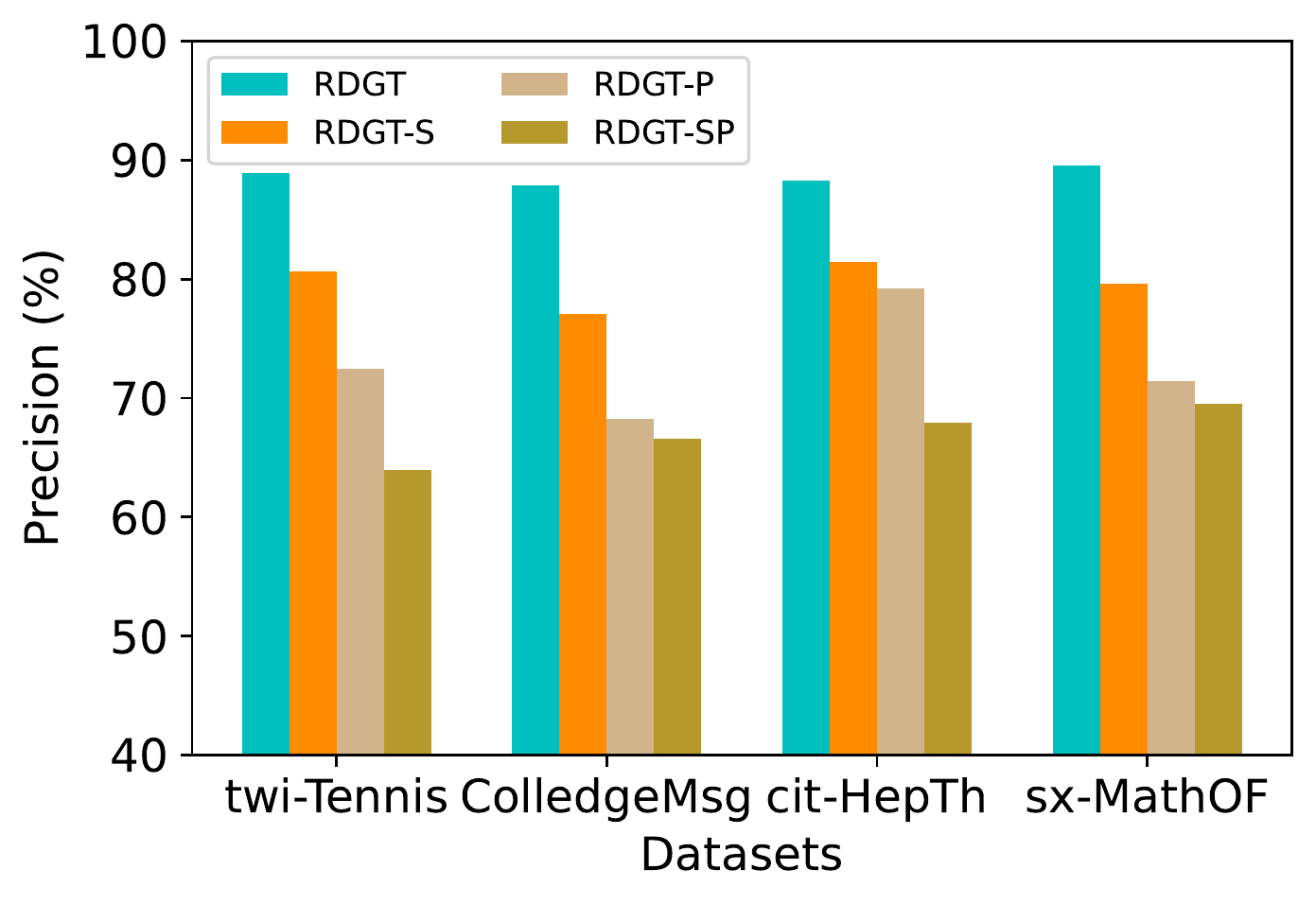}%
\label{fig:asp}}
\hfil
\subfloat[F1]{\includegraphics[width=0.24\textwidth]{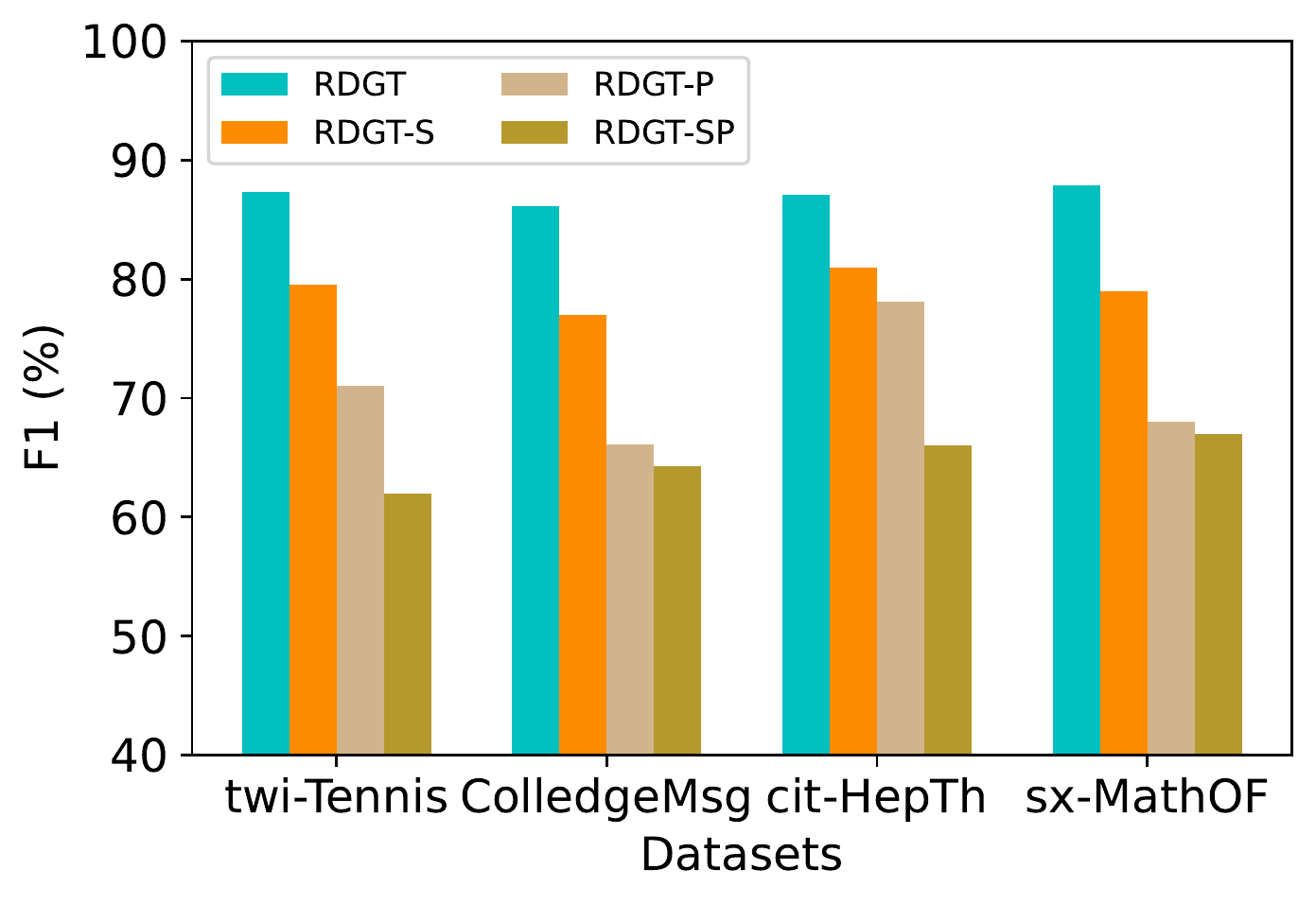}%
\label{fig:asf}}
\caption{Performance of different ablated variations of RSGT. (a)-(d) show the impact of edge temporal state modeling, while (e)-(h) illustrate the effect of graph topology learning across different performance metrics and datasets.}
\label{fig:ablation}
\end{figure*}

\subsection{Ablation Study}
To evaluate the individual contributions of each component within the RSGT, we performed a comprehensive ablation study on dynamic link prediction tasks across four datasets. Our ablation variants can be categorized into two main groups: those related to edge temporal state modeling and those related to graph topology learning.

\subsubsection{Edge Temporal State Modeling}
We first examine the impact of edge temporal state modeling through the following variants:
\begin{itemize}
    \item RSGT-T: Retains only edge weights without considering edge types. This variant helps us understand the importance of modeling short-term dynamics through edge type changes.
    \item RSGT-W: Retains only edge types without explicit edge weights modeling. This allows us to assess the impact of capturing long-term dynamics through the evolution of edge weights over time.
    \item RSGT-TW: Uses the original dynamic graph directly without explicit modeling of edge temporal states or weights. This variant serves as a baseline to evaluate the overall impact of our temporal modeling approach.
\end{itemize}

Figures \ref{fig:ac}-\ref{fig:af} illustrate the performance of these variants across all datasets. We observe that RSGT-T and RSGT-W perform similarly on the twi-Tennis, ColledgeMsg, and sx-MathOF datasets, with only minor differences in their performance metrics. This suggests that both edge types and weights contribute substantially to capturing temporal state information. The full RSGT model consistently outperforms these variants across all datasets, highlighting the importance of combining both edge types and weights.

Interestingly, on the cit-HepTh dataset, RSGT and RSGT-T demonstrate clear performance superiority over RSGT-W and RSGT-TW. This can be attributed to the unique characteristics of the cit-HepTh dataset, which is a citation network. In such networks, edges (citations) are inherently stable once formed, rarely disappearing over time. This stability emphasizes the importance of long-term dynamics in the graph structure.

\subsubsection{Graph Topology Learning}
To assess the importance of graph topology learning, we evaluate the following variants:
\begin{itemize}
    \item RSGT-S: In this variant, the feature learning process does not account for the topological attributes between node pairs such as the out-degree, in-degree, and shortest path length of nodes, thus not including $\textbf{attr}_{s}^t$.
    \item RSGT-P: In this variant, the feature learning process does not consider the shortest path information between node pairs, thus not including $\textbf{ATTR}_{p}^t$. This leads the model to overlook the explicit representation of temporal status information of edges.
    \item RSGT-SP: This variant does not integrate the complete topological structural information of the graph into the node representation learning process, rather it solely depends on the raw self-attention mechanism for learning the node representations.
\end{itemize}

When it comes to learning features of graph topology, as illustrated in Figure \ref{fig:asc}-\ref{fig:asf}, both RSGT-S and RSGT-P show significant performance deterioration across four all experimental datasets. Remarkably, the performance decline of RSGT-P is more severe than RSGT-S. This finding suggests that the shortest path, along with varying edge weights and types along the path, can more effectively clarify the topological dependency relationship between nodes.
As anticipated, RSGT-SP performed the worst among all ablated models due to its neglect of graph topology information, depending solely on the original self-attention mechanism to learn global semantic relevance between nodes, which fails to capture the topological dependencies between nodes. This further confirms the efficacy of our structure-reinforced attention mechanism that integrates graph topological information into the Transformer architecture.

In conclusion, RSGT outperforms all other ablation variants across all datasets, underscoring the importance and effectiveness of explicitly modeling the edge temporal status using different edge types and weights, and considering the graph's topology during the model learning process.

\begin{figure*}[!t]
\centering
\subfloat[]{\includegraphics[width=0.23\textwidth]{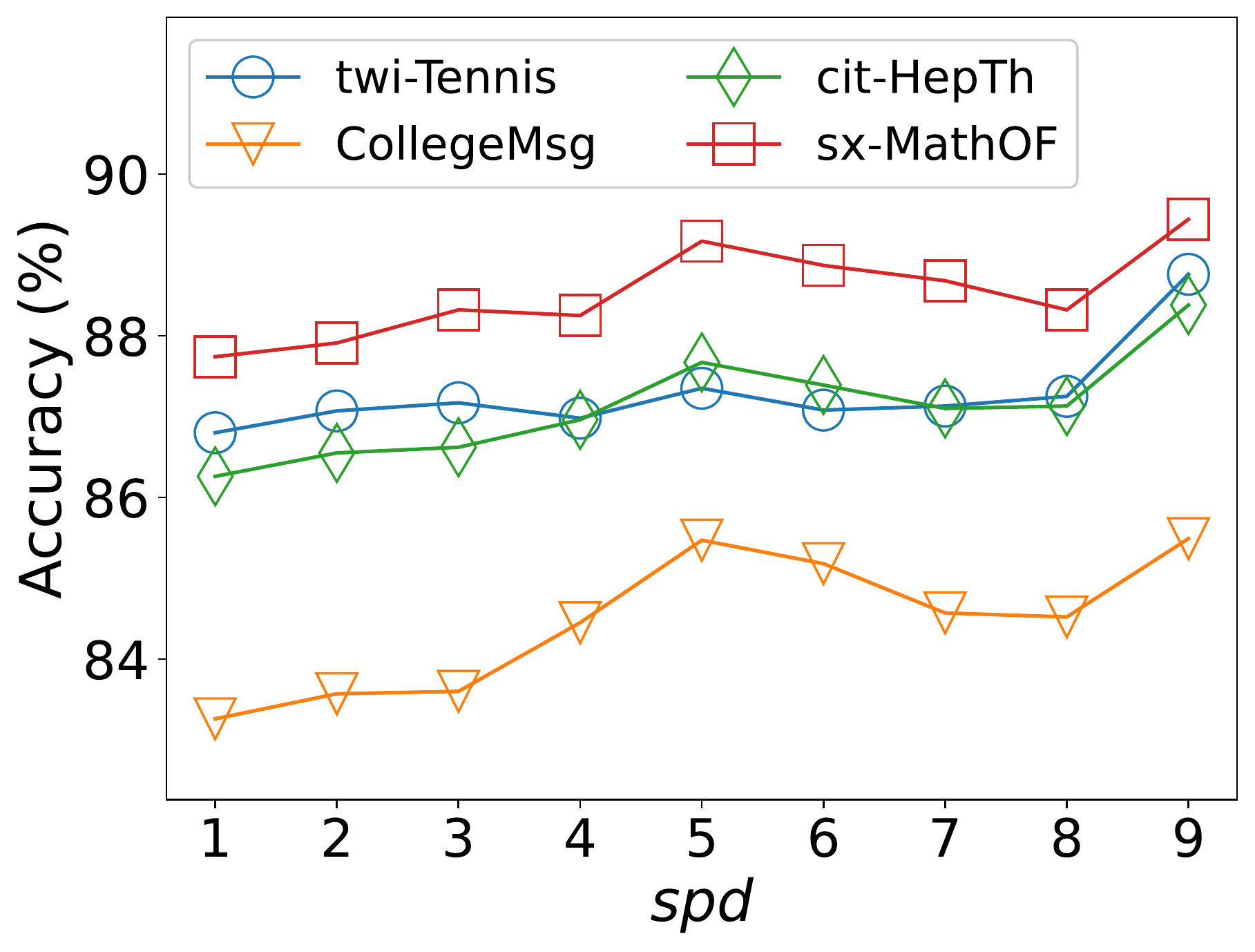}%
\label{fig:impact_a}}
\hfil
\subfloat[]{\includegraphics[width=0.23\textwidth]{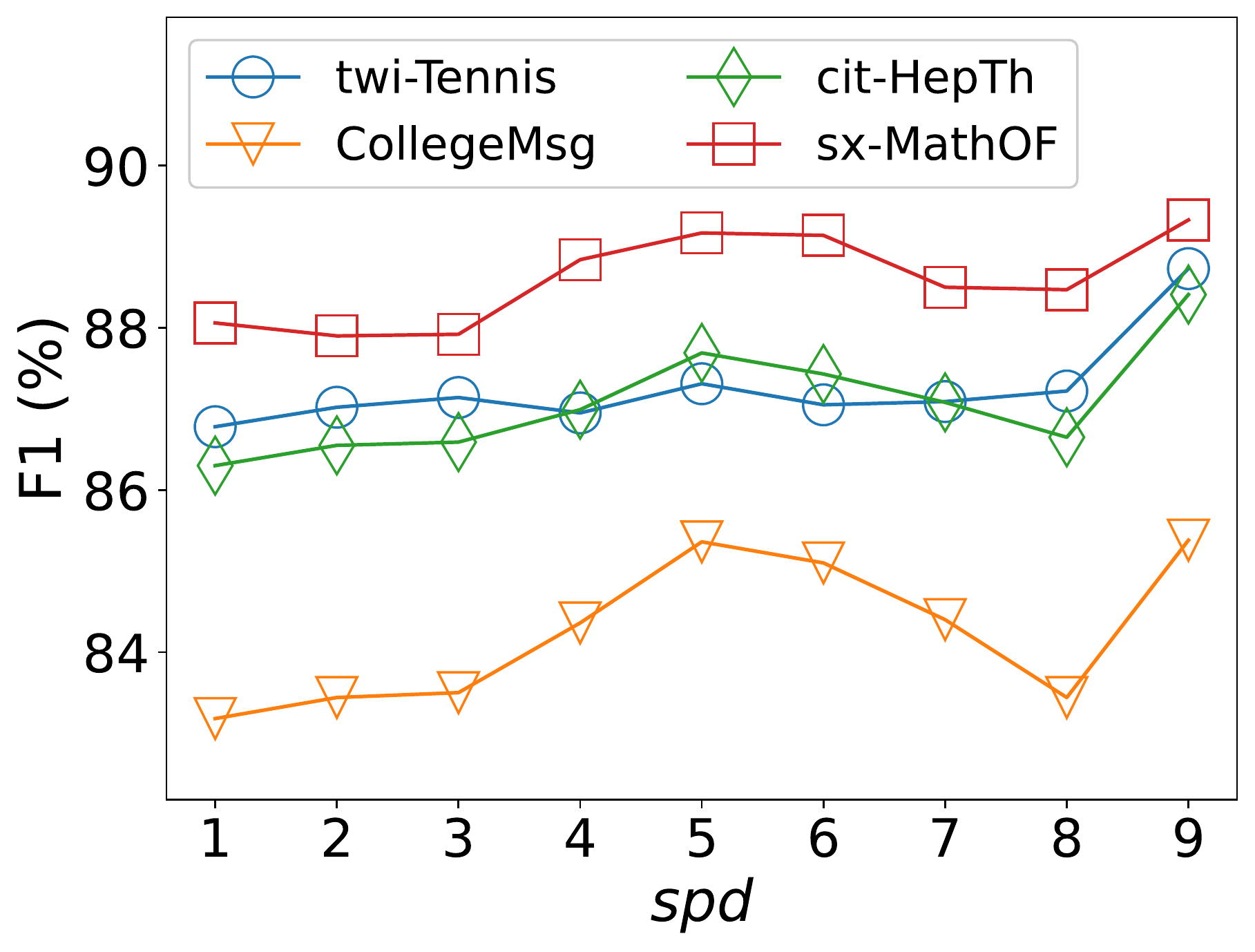}%
\label{fig:impact_b}}
\hfil
\subfloat[]{\includegraphics[width=0.23\textwidth]{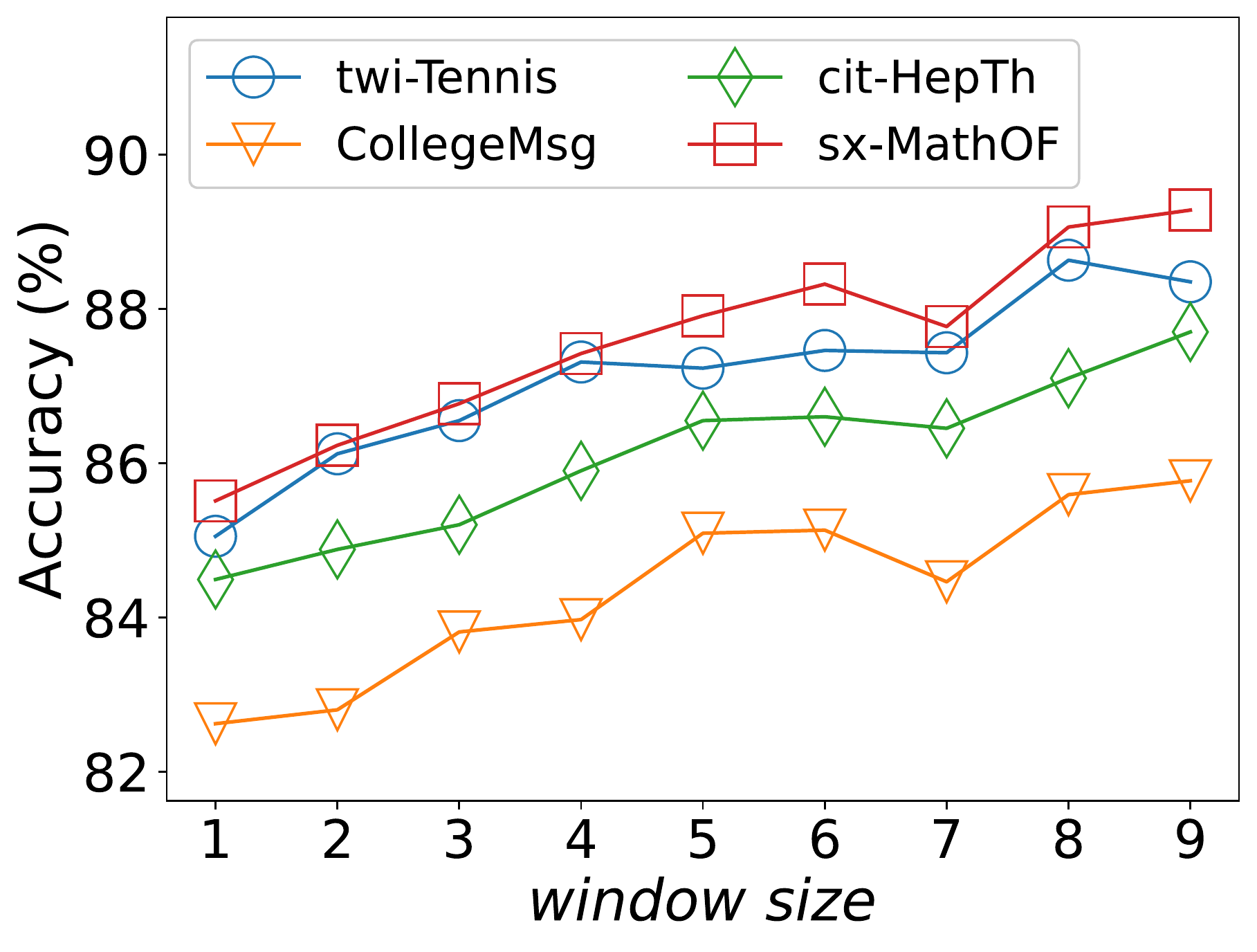}%
\label{fig:impact_c}}
\hfil
\subfloat[]{\includegraphics[width=0.23\textwidth]{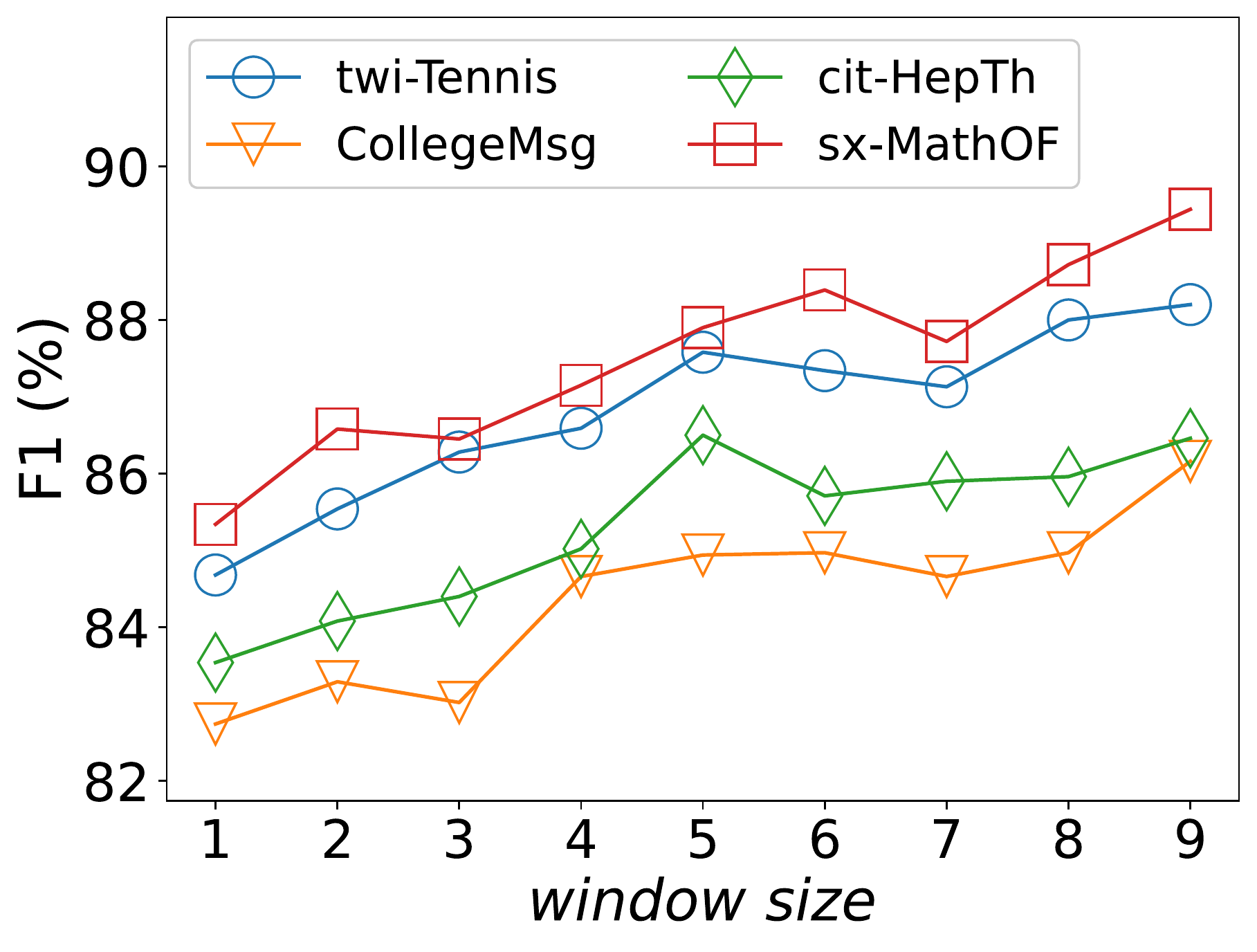}%
\label{fig:impact_d}}
\hfil
\subfloat[]{\includegraphics[width=0.23\textwidth]{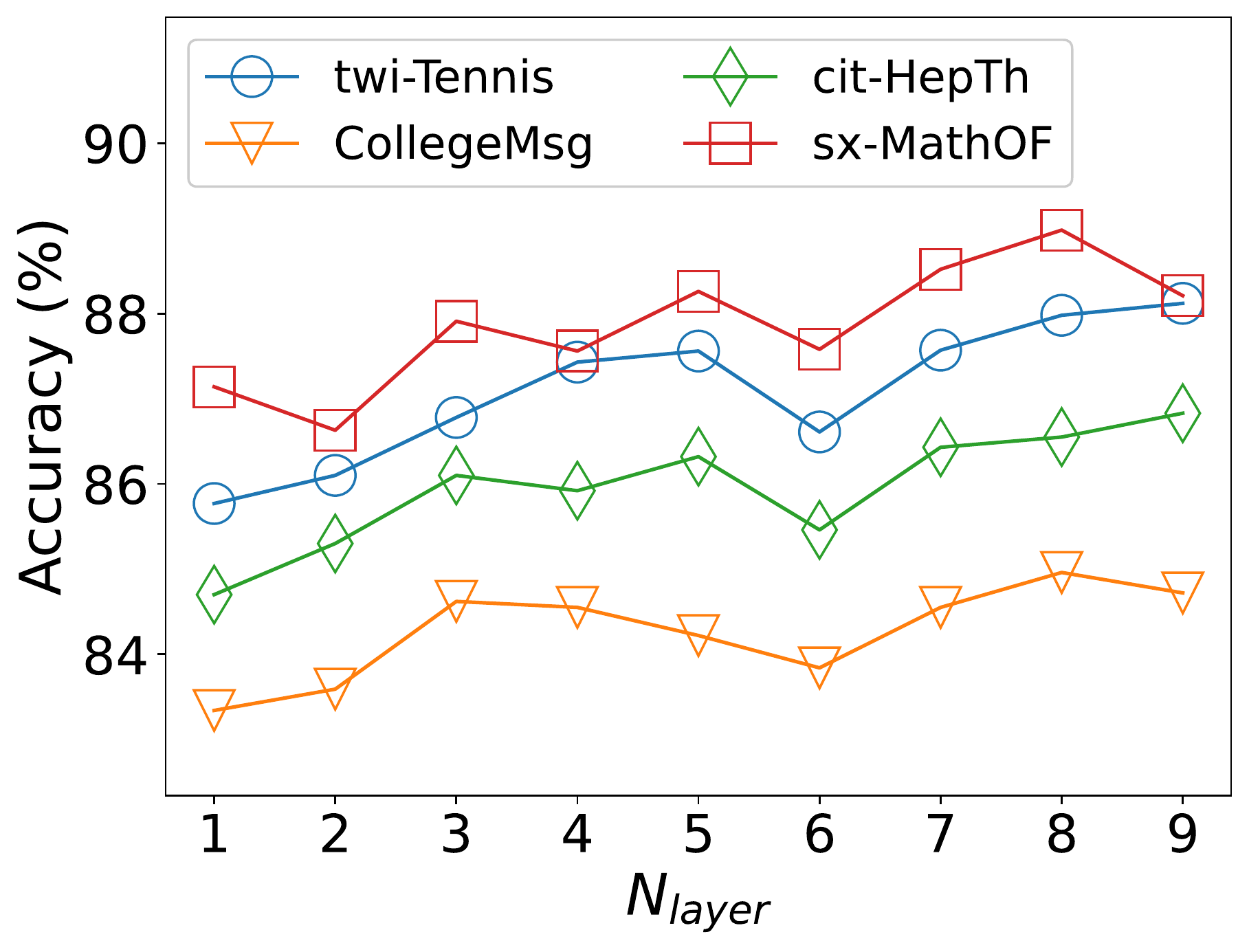}%
\label{fig:impact_e}}
\hfil
\subfloat[]{\includegraphics[width=0.23\textwidth]{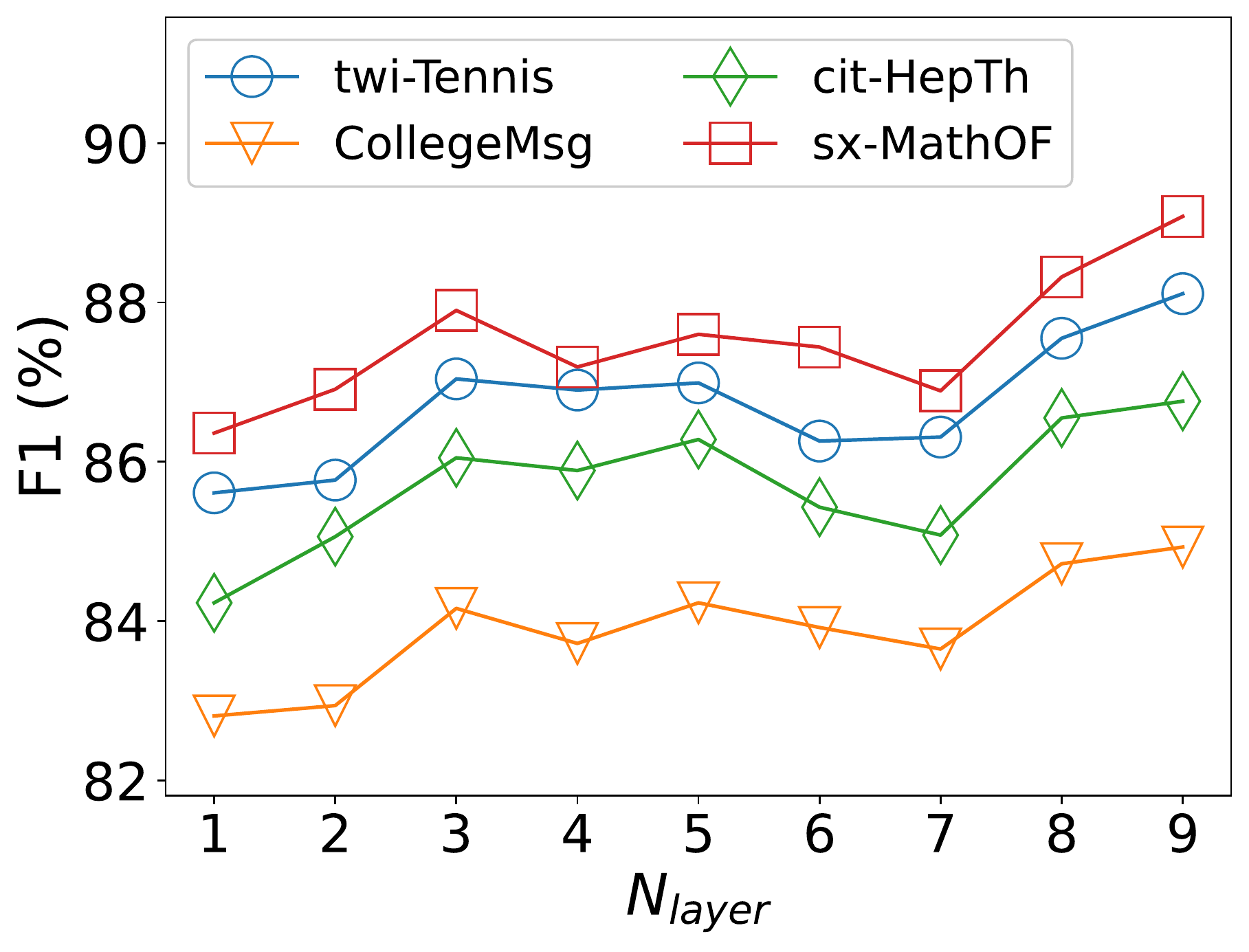}%
\label{fig:impact_f}}
\hfil
\subfloat[]{\includegraphics[width=0.23\textwidth]{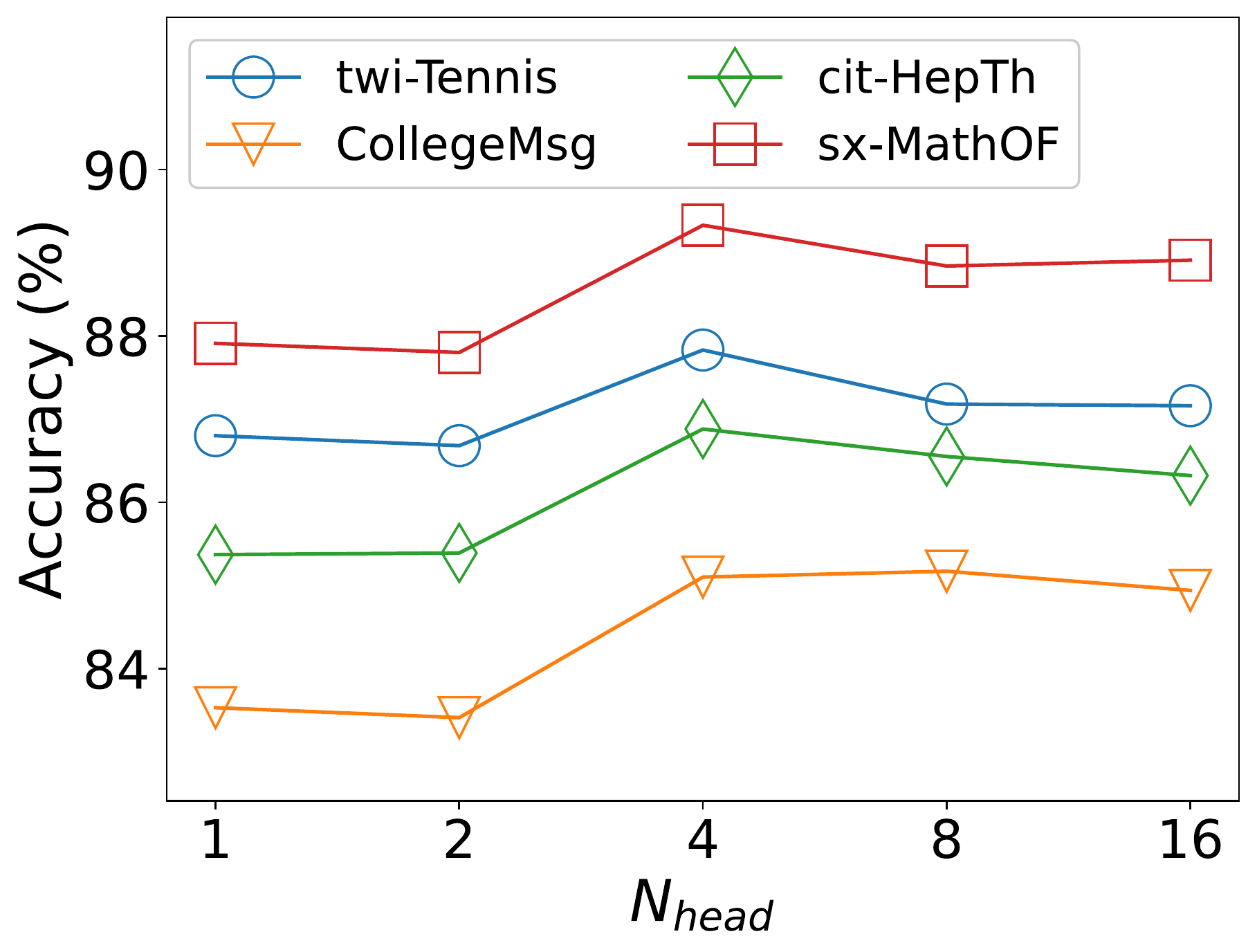}%
\label{fig:impact_g}}
\hfil
\subfloat[]{\includegraphics[width=0.23\textwidth]{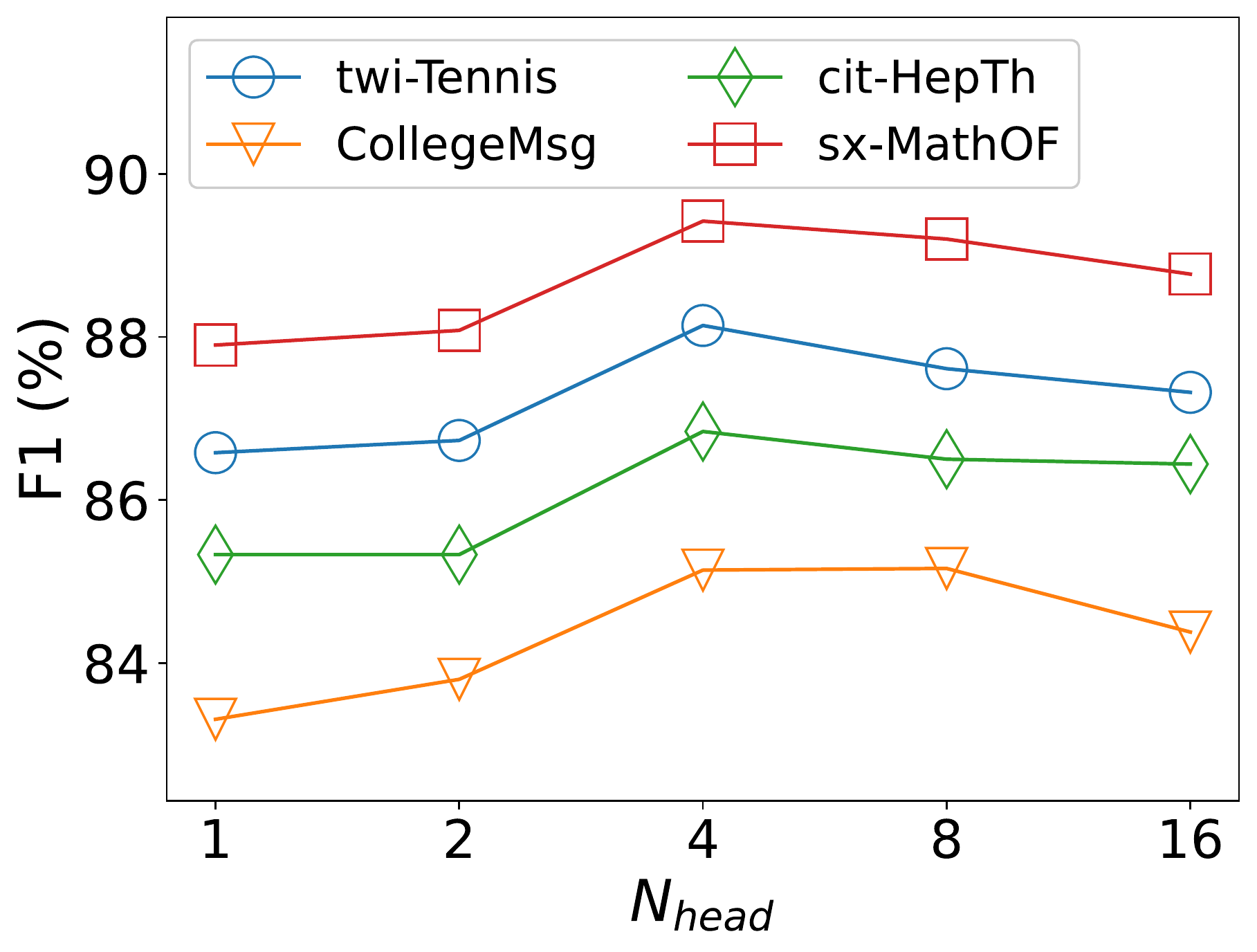}%
\label{fig:impact_h}}
\caption{Impact of key hyperparameters on RSGT performance across four datasets (twi-Tennis, CollegeMsg, cit-HepTh, sx-MathOF). The subplots show the effect of (a-b) shortest path distance ($spd$), (c-d) window size, (e-f) number of encoding layers ($N_{layer}$), and (g-h) number of attention heads ($N_{head}$) on model performance. The x-axis in each subplot represents the varying values of the respective parameter, and the y-axis shows the corresponding performance metric in percentage.}
\label{fig:impact}
\end{figure*}

\subsection{Performance Impact of Parameters}
To optimize the performance of our proposed RSGT model, we conducted a comprehensive analysis of four key hyperparameters: shortest path distance ($spd$), window size, number of encoding layers ($N_{layer}$), and number of attention heads ($N_{head}$). This analysis aims to understand the impact of each parameter on model performance and determine their optimal values for our experiments.

\textbf{\textit{Shortest Path Distance}} 
The $spd$ parameter determines the subgraph size used to learn the representation of the target node. We varied $spd$ from 1 to 9 and evaluated its impact on model performance.
As shown in Figures \ref{fig:impact_a} and \ref{fig:impact_b}, there is a positive correlation between $spd$ and model performance across all datasets. This trend suggests that our structure-reinforced graph transformer effectively leverages information from a broader context to learn more comprehensive and informative node representations.
The performance improvement is particularly notable for the sx-MathOF dataset, which may be due to its more complex network structure requiring a larger neighborhood for effective representation learning. However, we observe that the improvement tends to plateau after $spd=5$ for most datasets, likely due to the inclusion of less relevant information from distant nodes.
Based on these results and considering computational efficiency, we set $spd$ to 5 for twi-Tennis, CollegeMsg, and cit-HepTh datasets, and 2 for the sx-MathOF dataset in our comparative experiments.

\textbf{\textit{Window Size}}
The window size parameter defines the number of consecutive snapshots preceding the target time slice used for model training. We experimented with window sizes ranging from 1 to 9.
Figures \ref{fig:impact_c} and \ref{fig:impact_d} demonstrate that model performance generally improves as the window size increases. This trend indicates that our recurrent learning framework effectively utilizes long-term temporal data and autonomously extracts valuable historical information.
The improvement is most pronounced for the cit-HepTh dataset, which aligns with the nature of citation networks where long-term dependencies are crucial. Conversely, the CollegeMsg dataset shows less sensitivity to increased window size, possibly due to the more transient nature of messaging interactions.
To balance performance gains with computational costs, we set the window size to 5 for all datasets in our experiments.

\textbf{\textit{Number of Encoding Layer}}
The $N_{layer}$ parameter determines the depth of the structure-reinforced graph transformer and impacts its ability to learn deep latent features. We varied $N_{layer}$ from 1 to 9 in our experiments.
Figures \ref{fig:impact_e} and \ref{fig:impact_f} reveal a general trend of improved performance with increased $N_{layer}$. This suggests that deeper models can capture more complex patterns in the dynamic graphs.
However, we observe diminishing returns and even slight performance degradation for some datasets (e.g., cit-HepTh) at higher $N_{layer}$ values. This could be attributed to overfitting or the vanishing gradient problem in very deep networks.
Considering the trade-off between model performance and computational complexity, we chose $N_{layer}=4$ for our experiments, which provides a good balance across all datasets.

\textbf{\textit{Number of Attention Head}}
Multi-head attention allows the model to focus on different representation subspaces simultaneously. We experimented with $N_{head}$ values of 1, 2, 4, 8, and 16.
As shown in Figures \ref{fig:impact_g} and \ref{fig:impact_h}, increasing $N_{head}$ generally improves model performance, particularly from 1 to 4 heads. This aligns with the findings of \cite{vaswani2017attention}, demonstrating the benefit of attending to multiple representation subspaces.
However, we observe that performance tends to plateau or even slightly decrease beyond 4 heads for most datasets. This could be due to increased model complexity leading to overfitting, especially on smaller datasets.
Based on these results, we set $N_{head}=4$ in our experiments, which provides optimal performance across datasets while maintaining computational efficiency.

%% file: related_work.tex
\section{Related Work}
\label{sec:related}

Dynamic graph representation learning methods aim to incorporate temporal data during the process of node representation learning. Such methods strive to encapsulate both the structural characteristics and temporal dynamics of a graph. The categorization of these techniques is often dependent on the modeling strategy employed by the dynamic graph. For instance, methods such as \cite{seo2018structured,hajiramezanali2019variational,sankar2020dysat,pareja2020evolvegcn,you2022roland} are designed specifically for snapshot-based discrete dynamic graphs. In contrast, others, including \cite{nguyen2018continuous,DBLP:conf/iclr/XuRKKA20,DBLP:conf/iclr/WangCLL021,wen2022trend}, are developed for timestamp-based continuous dynamic graphs. 
This section offers a brief overview of the related works, which are divided into discrete dynamic graph representation learning and continuous dynamic graph representation learning.

\subsection{Discrete Dynamic Graph Representation Learning}
Building on the advancements in message-passing-based Graph Neural Networks (GNNs) and their state-of-the-art (SOTA) performance in various static graph tasks \cite{hamilton2017inductive}, recent research has focused on integrating GNNs with sequence models for the learning of discrete dynamic graph representations. In this setup, GNNs are primarily used to retrieve graph structural information, while sequence models are employed to capture the evolutionary dynamics.

Typically, structural features are initially extracted from each time slice's snapshot using GNNs. Subsequently, a sequence model captures the evolutionary dynamics between these snapshots and updates the features accordingly. For instance, Graph Convolutional Recurrent Network (GCRN) \cite{seo2018structured} utilizes a Graph Convolutional Network (GCN) \cite{hamilton2017inductive} to obtain node embeddings, which are then processed by a Long Short-Term Memory (LSTM) network to learn the evolutionary dynamics.
Variational Recurrent Graph Neural Network (VRGNN) \cite{hajiramezanali2019variational} extends GCRN by incorporating Variational Graph Auto-Encoder (VGAE) \cite{kipf2016variational} to enhance the expressive power of GCRN and to capture the uncertainty in node latent representation. DySAT \cite{sankar2020dysat} follows a similar approach to GCRN, but calculates node representations through a joint self-attention mechanism along the structural neighborhood and temporal dynamics dimensions.

Despite the intuitive nature and potential of these approaches for dynamic graph representation learning, they suffer from a lack of inductive learning capability, which is the inability to leverage the temporal information of neighboring nodes for newly introduced nodes. To address this limitation, several studies have explored methods to simultaneously learn both graph structural and dynamic features, thereby improving inductive learning. EvolveGCN \cite{pareja2020evolvegcn}, for example, proposes two architectures that capture the dynamism of the graph sequence by using a Recurrent Neural Network (RNN) to evolve the GCN parameters instead of the node embeddings.
As the number of nodes in a dynamic graph grows across time slices, DynGEM \cite{goyal2018dyngem} employs a deep autoencoder to progressively construct the representations of a particular snapshot from the embedding of the preceding snapshot. CoEvoSAGE \cite{wang2021modeling} leverages GraphSAGE to learn the structural dependencies for each snapshot and introduces a temporal self-attention architecture to model long-range dependencies.
ROLAND \cite{you2022roland} is designed to easily adapt any static GNN for use with dynamic graphs. It treats the node embeddings at different GNN layers as hierarchical node states and updates them recurrently over time.
DynGNN \cite{9858109} integrates an RNN with a GNN for effectively capturing the evolving patterns of graphs by considering both temporal dynamics and graph topology.

\subsection{Continuous Dynamic Graph Representation Learning}
The continuous dynamic graph models edges at different temporal instances as timestamped events. These can be represented as a set of triplets $\{\langle v_i, v_j, t\rangle\}$, where $\langle v_i, v_j, t\rangle$ signifies the presence of an edge between nodes $v_i$ and $v_j$ at the timestamp $t$. Various representation learning methods for continuous dynamic graphs \cite{nguyen2018continuous,DBLP:conf/iclr/XuRKKA20,DBLP:conf/iclr/WangCLL021,wen2022trend} seek to understand and elucidate the evolving patterns of these graphs from such temporal event sequences.

Continuous dynamic graph representation learning methods can be broadly categorized into two main approaches: Sequence-based modeling methods, and Advanced temporal modeling methods.
Sequence-based modeling methods aim to capture temporal-topological information by treating the graph evolution as a sequence of events. CTDNE \cite{nguyen2018continuous} generalizes random walk-based embedding techniques to continuous dynamic graphs, effectively capturing local temporal patterns but potentially struggling with long-range dependencies. Attention-based methods like TGAT \cite{DBLP:conf/iclr/XuRKKA20} and CAW \cite{DBLP:conf/iclr/WangCLL021} leverage self-attention mechanisms to capture temporal-topological interactions. TGAT introduces a temporal graph attention layer to aggregate features from temporal-topological neighborhoods, while CAW proposes Causal Anonymous Walks as an automatic retrieval mechanism for temporal graphs. These methods excel at capturing complex temporal dependencies but may face computational challenges with very large graphs. DyGFormer \cite{yu2023towards} employs a Transformer-based architecture with neighbor co-occurrence encoding and patching techniques to learn from nodes' historical first-hop interactions, effectively capturing both short-term and long-term dependencies.

Advanced temporal modeling methods employ sophisticated techniques to model the continuous evolution of graphs. TREND \cite{wen2022trend} leverages Hawkes processes to incorporate the evolutionary characteristics of temporal edges into node representations. R-GSAGE \cite{yao2024recurrent} uses neural ordinary differential equations to model the continuous dynamic evolution of node embedding trajectories, allowing for a more nuanced representation of temporal dynamics. GraSSP \cite{celikkanat2024continuous} proposes a novel stochastic process to model link durations and their absences in continuous-time graphs. FuzzyDGL \cite{10286559} applies fuzzy logic to address uncertainties in dynamic graphs, enhancing representation quality by capturing structural and temporal correlations with improved feature fuzziness handling. ConTIG \cite{wang2024contig} integrates a recurrent structure into GraphSAGE to jointly explore structural and temporal patterns while maintaining a lightweight architecture, making it particularly efficient for large-scale dynamic graphs.

Despite these advancements, existing approaches still face several challenges. Many methods learn evolutionary patterns implicitly, without explicitly considering the various temporal states of edges and their impact on node representations. Furthermore, the prevalent use of GNNs as the backbone for structural feature extraction often leads to over-smoothing issues, limiting model performance. Additionally, most current approaches struggle to effectively balance the modeling of both short-term and long-term temporal dependencies, especially in graphs with diverse temporal scales.
Our proposed RSGT addresses these limitations by introducing a novel edge temporal state modeling technique and a structure-reinforced graph transformer. It allows for explicit modeling of edge dynamics while effectively capturing both local and global structural information, thereby overcoming the challenges faced by existing methods.

%% file: main.bbl
\begin{thebibliography}{10}
\expandafter\ifx\csname url\endcsname\relax
  \def\url#1{\texttt{#1}}\fi
\expandafter\ifx\csname urlprefix\endcsname\relax\def\urlprefix{URL }\fi
\expandafter\ifx\csname href\endcsname\relax
  \def\href#1#2{#2} \def\path#1{#1}\fi

\bibitem{10184800}
Y.~Zhao, X.~Luo, W.~Ju, C.~Chen, X.-S. Hua, M.~Zhang, {Dynamic Hypergraph Structure Learning for Traffic Flow Forecasting}, in: International Conference on Data Engineering (ICDE), 2023, pp. 2303--2316.

\bibitem{DBLP:journals/tits/DuCLCL24}
W.~Du, S.~Chen, Z.~Li, X.~Cao, Y.~Lv, {A Spatial-Temporal Approach for Multi-Airport Traffic Flow Prediction Through Causality Graphs}, {IEEE} Transactions on Intelligent Transportation Systems 25~(1) (2024) 532--544.

\bibitem{DBLP:journals/tois/YiOM24}
Z.~Yi, I.~Ounis, C.~MacDonald, {Contrastive Graph Prompt-tuning for Cross-domain Recommendation}, {ACM} Transactions on Information Systems 42~(2) (2024) 60:1--60:28.

\bibitem{DBLP:journals/tkde/YuXCCHY24}
J.~Yu, X.~Xia, T.~Chen, L.~Cui, N.~Q.~V. Hung, H.~Yin, {XSimGCL: Towards Extremely Simple Graph Contrastive Learning for Recommendation}, {IEEE} Transactions on Knowledge And Data Engineering 36~(2) (2024) 913--926.

\bibitem{DBLP:journals/tkde/LiuWWYDZW24}
C.~Liu, W.~Wu, S.~Wu, L.~Yuan, R.~Ding, F.~Zhou, Q.~Wu, {Social-Enhanced Explainable Recommendation With Knowledge Graph}, {IEEE} Transactions on Knowledge And Data Engineering 36~(2) (2024) 840--853.

\bibitem{10176355}
H.~Tian, X.~Zhang, X.~Zheng, D.~D. Zeng, {Learning Dynamic Dependencies With Graph Evolution Recurrent Unit for Stock Predictions}, IEEE Transactions on Systems, Man, and Cybernetics: Systems 53~(11) (2023) 6705--6717.

\bibitem{pareja2020evolvegcn}
A.~Pareja, G.~Domeniconi, J.~Chen, T.~Ma, T.~Suzumura, H.~Kanezashi, T.~Kaler, T.~Schardl, C.~Leiserson, {EvolveGCN: Evolving Graph Convolutional Networks for Dynamic Graphs}, in: AAAI Conference on Artificial Intelligence (AAAI), Vol.~34, 2020, pp. 5363--5370.

\bibitem{you2022roland}
J.~You, T.~Du, J.~Leskovec, {ROLAND: Graph Learning Framework for Dynamic Graphs}, in: ACM SIGKDD Conference on Knowledge Discovery and Data Mining (SIGKDD), 2022, pp. 2358--2366.

\bibitem{nguyen2018continuous}
G.~H. Nguyen, J.~B. Lee, R.~A. Rossi, N.~K. Ahmed, E.~Koh, S.~Kim, {Continuous-Time Dynamic Network Embeddings}, in: Companion Proceedings of the The Web Conference (WWW), 2018, p. 969–976.

\bibitem{wen2022trend}
Z.~Wen, Y.~Fang, {TREND: TempoRal Event and Node Dynamics for Graph Representation Learning}, in: ACM Web Conference (WWW), 2022, pp. 1159--1169.

\bibitem{perozzi2014deepwalk}
B.~Perozzi, R.~Al-Rfou, S.~Skiena, {Deepwalk: Online learning of social representations}, in: ACM SIGKDD International Conference on Knowledge Discovery and Data Mining (SIGKDD), 2014, pp. 701--710.

\bibitem{grover2016node2vec}
A.~Grover, J.~Leskovec, {Node2vec: Scalable Feature Learning for Networks}, in: ACM SIGKDD International Conference on Knowledge Discovery and Data Mining (SIGKDD), 2016, pp. 855--864.

\bibitem{wang2021modeling}
D.~Wang, Z.~Zhang, Y.~Ma, T.~Zhao, T.~Jiang, N.~Chawla, M.~Jiang, {Modeling Co-Evolution of Attributed and Structural Information in Graph Sequence}, IEEE Transactions on Knowledge and Data Engineering 35~(2) (2021) 1817--1830.

\bibitem{hu22temporal}
S.~Hu, G.~Zou, B.~Zhang, S.~Wu, S.~Lin, Y.~Gan, Y.~Chen, {Temporal-Aware QoS Prediction via Dynamic Graph Neural Collaborative Learning }, in: International Conference on Service-Oriented Computing (ICSOC), 2022, pp. 125--133.

\bibitem{DBLP:conf/iclr/XuRKKA20}
D.~Xu, C.~Ruan, E.~K{\"{o}}rpeoglu, S.~Kumar, K.~Achan, {Inductive Representation Learning on Temporal Graphs}, in: International Conference on Learning Representations (ICLR), 2020.

\bibitem{DBLP:conf/iclr/WangCLL021}
Y.~Wang, Y.~Chang, Y.~Liu, J.~Leskovec, P.~Li, {Inductive Representation Learning in Temporal Networks via Causal Anonymous Walks}, in: International Conference on Learning Representations (ICLR), 2021.

\bibitem{yao2024recurrent}
H.-Y. Yao, C.-Y. Zhang, Z.-L. Yao, C.~P. Chen, J.~Hu, {A Recurrent Graph Neural Network for Inductive Representation Learning on Dynamic Graphs}, Pattern Recognition 154 (2024) 110577.

\bibitem{celikkanat2024continuous}
A.~Celikkanat, N.~Nakis, M.~M{\o}rup, {Continuous-time Graph Representation with Sequential Survival Process}, in: Proceedings of the AAAI Conference on Artificial Intelligence (AAAI), Vol.~38, 2024, pp. 11177--11185.

\bibitem{goyal2018dyngem}
P.~Goyal, N.~Kamra, X.~He, Y.~Liu, {Dyngem: Deep embedding method for dynamic graphs}, arXiv preprint arXiv:1805.11273 (2018).

\bibitem{wang2024contig}
Z.~Wang, P.~Yang, X.~Fan, X.~Yan, Z.~Wu, S.~Pan, L.~Chen, Y.~Zang, C.~Wang, R.~Yu, {Contig: Continuous Representation Learning on Temporal Interaction Graphs}, Neural Networks 172 (2024) 106151.

\bibitem{zhu2023path}
H.~Zhu, X.~Yang, J.~Wei, {Path Prediction of Information Diffusion based on a Topic-oriented Relationship Strength Network}, Information Sciences 631 (2023) 108--119.

\bibitem{zhou2022point}
Y.~Zhou, G.~Yang, B.~Yan, Y.~Cai, Z.~Zhu, Point-of-interest recommendation model considering strength of user relationship for location-based social networks, Expert Systems with Applications 199 (2022) 117147.

\bibitem{ying2021transformers}
C.~Ying, T.~Cai, S.~Luo, S.~Zheng, G.~Ke, D.~He, Y.~Shen, T.-Y. Liu, {Do Transformers Really Perform Badly for Graph Representation?}, in: Advances in Neural Information Processing Systems (NeurIPS), Vol.~34, 2021, pp. 28877--28888.

\bibitem{chen2022structure}
D.~Chen, L.~O'Bray, K.~Borgwardt, {Structure-aware transformer for graph representation learning}, in: International Conference on Machine Learning (ICML), 2022, pp. 3469--3489.

\bibitem{chen2020measuring}
D.~Chen, Y.~Lin, W.~Li, P.~Li, J.~Zhou, X.~Sun, {Measuring and relieving the over-smoothing problem for graph neural networks from the topological view}, in: Proceedings of the AAAI conference on artificial intelligence, Vol.~34, 2020, pp. 3438--3445.

\bibitem{rusch2023survey}
T.~K. Rusch, M.~M. Bronstein, S.~Mishra, {A survey on oversmoothing in graph neural networks}, arXiv preprint arXiv:2303.10993 (2023).

\bibitem{quach2021dyglip}
K.~G. Quach, P.~Nguyen, H.~Le, T.-D. Truong, C.~N. Duong, M.-T. Tran, K.~Luu, {DyGLIP: A Dynamic Graph Model With Link Prediction for Accurate Multi-Camera Multiple Object Tracking}, in: IEEE/CVF Conference on Computer Vision and Pattern Recognition (CVPR), 2021, pp. 13784--13793.

\bibitem{xu2019spatio}
D.~Xu, W.~Cheng, D.~Luo, X.~Liu, X.~Zhang, {Spatio-Temporal Attentive RNN for Node Classification in Temporal Attributed Graphs}, in: International Joint Conference on Artificial Intelligence (IJCAI), 2019, pp. 3947--3953.

\bibitem{wang2019neural}
X.~Wang, X.~He, M.~Wang, F.~Feng, T.-S. Chua, {Neural graph collaborative filtering}, in: Proceedings of the 42nd international ACM SIGIR conference on Research and development in Information Retrieval, 2019, pp. 165--174.

\bibitem{vaswani2017attention}
A.~Vaswani, N.~Shazeer, N.~Parmar, Uszkoreit, A.~N. Gomez, L.~u. Kaiser, I.~Polosukhin, {Attention is All you Need}, in: Advances in Neural Information Processing Systems (NeurIPS), Vol.~30, 2017, pp. 5998--6008.

\bibitem{borgatti2005centrality}
S.~P. Borgatti, {Centrality and Network Flow}, Social Networks 27~(1) (2005) 55--71.

\bibitem{newman2018networks}
M.~Newman, {Networks}, 2018.

\bibitem{loshchilov2017decoupled}
I.~Loshchilov, F.~Hutter, {Decoupled weight decay regularization} (2019).

\bibitem{beres2018temporal}
F.~B{\'e}res, R.~P{\'a}lovics, A.~Ol{\'a}h, A.~A. Bencz{\'u}r, {Temporal Walk Based Centrality Metric for Graph Streams}, Applied Network Science 3~(1) (2018) 1--26.

\bibitem{2009patterns}
P.~Panzarasa, T.~Opsahl, K.~M. Carley, {Patterns and dynamics of users' behavior and interaction: Network analysis of an online community}, Journal of the American Society for Information Science and Technology 60~(5) (2009) 911--932.

\bibitem{2005grpahsovertime}
J.~Leskovec, J.~Kleinberg, C.~Faloutsos, {Graphs over Time: Densification Laws, Shrinking Diameters and Possible Explanations}, in: ACM SIGKDD International Conference on Knowledge Discovery and Data Mining (SIGKDD), 2005, p. 177–187.

\bibitem{paranjape2017motifs}
A.~Paranjape, A.~R. Benson, J.~Leskovec, {Motifs in Temporal Networks}, in: ACM International Conference on Web Search and Data Mining (WSDM), 2017, pp. 601--610.

\bibitem{mikolov2013distributed}
T.~Mikolov, I.~Sutskever, K.~Chen, G.~S. Corrado, J.~Dean, {Distributed Representations of Words and Phrases and their Compositionality}, in: Advances in Neural Information Processing Systems (NeurIPS), Vol.~26, 2013, pp. 3111--3119.

\bibitem{hamilton2017inductive}
W.~Hamilton, Z.~Ying, J.~Leskovec, {Inductive Representation Learning on Large Graphs}, in: Advances in Neural Information Processing Systems (NeurIPS), Vol.~30, 2017, pp. 1024--1034.

\bibitem{hawkes1971spectra}
A.~G. Hawkes, {Spectra of Some Self-Exciting and Mutually Exciting Point Processes}, Biometrika 58~(1) (1971) 83--90.

\bibitem{yu2023towards}
L.~Yu, L.~Sun, B.~Du, W.~Lv, {Towards Better Dynamic Graph Learning: New Architecture and Unified Library}, in: Advances in Neural Information Processing Systems (NeurIPS), 2023, pp. 67686--67700.

\bibitem{DBLP:conf/icml/SarkarCJ12}
P.~Sarkar, D.~Chakrabarti, M.~I. Jordan, {Nonparametric Link Prediction in Dynamic Networks}, in: International Conference on Machine Learning (ICML), 2012.

\bibitem{seo2018structured}
Y.~Seo, M.~Defferrard, P.~Vandergheynst, X.~Bresson, {Structured Sequence Modeling with Graph Convolutional Recurrent Networks}, in: International Conference on Neural Information Processing (ICONIP), 2018, pp. 362--373.

\bibitem{hajiramezanali2019variational}
E.~Hajiramezanali, A.~Hasanzadeh, K.~Narayanan, N.~Duffield, M.~Zhou, X.~Qian, {Variational Graph Recurrent Neural Networks}, in: Advances in Neural Information Processing Systems (NeurIPS), Vol.~32, 2019, pp. 10700--10710.

\bibitem{sankar2020dysat}
A.~Sankar, Y.~Wu, L.~Gou, W.~Zhang, H.~Yang, {DySAT: Deep Neural Representation Learning on Dynamic Graphs via Self-Attention Networks}, in: International Conference on Web Search and Data Mining (WSDM), 2020, pp. 519--527.

\bibitem{kipf2016variational}
T.~N. Kipf, M.~Welling, {Variational graph auto-encoders}, arXiv preprint arXiv:1611.07308 (2016).

\bibitem{9858109}
C.~Zhang, Z.~Yao, H.~Yao, Huang, {Dynamic Representation Learning via Recurrent Graph Neural Networks}, IEEE Transactions on Systems, Man, and Cybernetics: Systems 53~(2) (2023) 1284--1297.

\bibitem{10286559}
H.~Yao, Y.~Yu, C.~Zhang, L.~Yuena, S.~Li, {Fuzzy Representation Learning on Dynamic Graphs}, IEEE Transactions on Systems, Man, and Cybernetics: Systems 54~(2) (2024) 878--890.

\end{thebibliography}
